\documentclass{article}

\PassOptionsToPackage{numbers, compress}{natbib}

\usepackage[final]{neurips_2021}

\usepackage[utf8]{inputenc} 
\usepackage[T1]{fontenc}    
\usepackage[dvipsnames,table]{xcolor}         
\usepackage[colorlinks=true,linkcolor=red,citecolor=blue]{hyperref}%
\usepackage{url}            
\usepackage{booktabs}       
\usepackage{amsfonts}       
\usepackage{nicefrac}       
\usepackage{microtype}      
\usepackage{amsmath}
\usepackage{graphicx}
\usepackage{tabularx}
\usepackage{bbm}
\usepackage{array}
\usepackage{arydshln}
\usepackage{amsfonts}
\usepackage{amsmath}

\usepackage{bbm}
\usepackage{boldline}
\usepackage{bigstrut}
\usepackage{blindtext}
\usepackage{booktabs, siunitx}
\usepackage[labelfont=bf, format=plain, justification=justified, singlelinecheck=false]{caption}
\usepackage{color}
\usepackage{cprotect}
\usepackage{ctable}
\usepackage{dirtytalk}
\usepackage{enumitem}
\usepackage[export]{adjustbox}
\usepackage{float}
\usepackage{graphicx}
\usepackage{hhline}
\usepackage{latexsym}
\usepackage{mathrsfs}
\usepackage{mathtools}
\usepackage{microtype}
\usepackage{moresize}
\usepackage{multicol}
\usepackage{multirow}
\usepackage{nccmath}
\usepackage{nicefrac}
\usepackage{pifont}
\usepackage{placeins}
\usepackage{soul}
\usepackage{subcaption}
\usepackage{times}
\usepackage[utf8]{inputenc}
\usepackage{url}
\usepackage{verbatim}
\usepackage{wrapfig, lipsum}
\usepackage{textcomp}
\usepackage{enumitem}
\usepackage{colortbl}
\usepackage[ruled,vlined]{algorithm2e}

\DeclarePairedDelimiterX{\infdivx}[2]{(}{)}{%
  #1\;\delimsize|\delimsize|\;#2%
}
\newcommand{\kld}[2]{\ensuremath{\mathbb{KL}\infdivx{#1}{#2}}\xspace}
\newcommand{\tabitem}{~~\llap{\textbullet}~~}

\newcommand{\cut}[1]{}

\newcommand{\cmark}{\ding{51}}

\newcommand{\vect}[1]{\mathbf{#1}}
\newcommand{\set}[1]{\mathcal{#1}}
\newcommand{\emdr}[1]{\textsc{Emdr}$^2$}

\setcitestyle{authoryear}
\setcitestyle{round}

\title{End-to-End Training of Multi-Document Reader and Retriever for Open-Domain Question Answering}

\author{Devendra Singh Sachan$^{1,2}$, Siva Reddy$^{1,2}$, William Hamilton$^{1,2}$, Chris Dyer$^{3}$, Dani Yogatama$^{3}$ \\
$^{1}$Mila - Quebec AI Institute\\
$^{2}$School of Computer Science, McGill University\\
$^{3}$DeepMind\\
{\tt {sachande@mila.quebec}, \{siva, wlh\}@cs.mcgill.ca }\\
{\tt \{cdyer, dyogatama\}@deepmind.com}
}

\begin{document}

\maketitle


\begin{abstract}
We present an end-to-end differentiable training method for 
retrieval-augmented open-domain question answering systems
that combine information from multiple retrieved documents when
generating answers. 
We model retrieval decisions as latent variables over sets of relevant documents.
Since marginalizing over sets of retrieved documents is computationally hard, 
we approximate this using an expectation-maximization algorithm.
We iteratively estimate the value of our latent variable (the set of relevant documents for a given question) and then use this estimate to update the retriever and reader parameters. 
We hypothesize that such end-to-end training allows training signals to flow to the reader and then to the retriever better than stage-wise training.
This results in a retriever that is able to select more relevant documents for a question and a reader that is trained on more accurate documents to generate an answer.
Experiments on three benchmark datasets demonstrate that our proposed method outperforms all existing approaches of comparable size by 2-3 absolute exact match points, achieving new state-of-the-art results.
Our results also demonstrate the feasibility of learning to retrieve to improve answer generation without explicit supervision of retrieval decisions.
\end{abstract}


\section{Introduction} 
\label{sec:introduction}
Open-domain question answering (OpenQA) is a question answering task where
the goal is to train a language model to produce an answer for a given question.
In contrast to many question answering tasks, an OpenQA model is only provided with 
the question as its input without accompanying documents that contain the answer.
One of the most promising approaches to OpenQA is based on
augmenting the language model with an external knowledge source such as Wikipedia
(often referred to as the evidence documents).
In this approach, the model consists of two core components~\citep{chen-etal-2017-reading}: (i) an information retrieval system to identify 
useful pieces of text from the knowledge source (the retriever); and (ii) a system to produce the answer given the retrieved documents and the question (the reader).

We can view such a model as a latent variable model, where the latent variables represent retrieved documents that are used to produce answers given questions~\citep{lee-etal-2019-latent}.
End-to-end (joint) training of this model is challenging since we need to learn both to generate an answer given retrieved documents and what to retrieve.
Previous work considers two potential solutions (see Table~\ref{tab:cmp} for a high-level summary).
First, they adopt a stage-wise training, where the retriever is trained while freezing the reader and vice versa \citep{karpukhin2020dense, izacard2021leveraging, izacard2021distilling}.
Another alternative is to constraint the reader to condition on each retrieved document individually\footnote{This makes marginalization over the latent variables easier since we only need to consider one document at a time rather 
than multiple documents at once.}
\citep{guu2020realm}---sometimes with extra supervision for the latent variables in the form of the relevant document for a question \citep{lewis20rag}.

In this paper, we consider a retrieval-augmented question answering model that combines information from multiple documents when generating answers.
Expectation-maximization \citep{em} offers a principled template for learning this class of latent variable models.
We present \emdr{}: \textbf{E}nd-to-end training of \textbf{M}ulti-\textbf{D}ocument \textbf{R}eader and \textbf{R}etriever (\S{\ref{sec:model}}).
\emdr{} iteratively uses feedback from the model itself as ``pseudo labels'' of the latent variables for optimizing the retriever and reader parameters.
We use two estimates of the latent variables: (i) prior scores for updating the reader parameters and (ii) approximate posterior scores given all observed variables for the retriever parameters.

We evaluate our proposed method by experimenting on three commonly used OpenQA datasets: Natural Questions, TriviaQA, and WebQuestions (\S\ref{sec:exp-setup}). 
\emdr{} achieves new state-of-the-art results for models of comparable size on all datasets, outperforming recent approaches by 2-3 absolute exact match points. 
We also show that \emdr{} is robust to retriever initialization.
It achieves high accuracy with unsupervised initialization,
suggesting that supervised training of the retriever 
may not be an essential component of the training process as suggested in prior work~\citep{karpukhin2020dense}. 

In summary, our contributions are as follows:
(i) we present an end-to-end training method (\emdr{}) for retrieval-augmented question-answering systems;
(ii) we demonstrate that \emdr{} outperforms other existing approaches of comparable size
without any kind of supervision on the latent variables;
(iii) we provide ablation studies for a better understanding of
the contributions of different components of our proposed method; and
(iv) we release our code and checkpoints to facilitate future work and for reproducibility.\footnote{Our code is available at: \url{https://github.com/DevSinghSachan/emdr2}}

\emdr{} is a framework that can be used to train retrieval-augmented text generation models for any task. We believe that our estimation technique in \emdr{} is also useful for learning similar latent variable models in other domains.

\begin{table*}[t]
\small
\centering
\setlength{\tabcolsep}{0.30em}
\begin{tabular}{@{}l c c c c c c@{}}
 \toprule
 & & & \multicolumn{4}{c}{\textbf{Reader and Retriever Training}} \\
 \midrule
 \textbf{Model} & \multirow{2}{1.4cm}{\textit{Multi-Doc Reader}} & \multirow{2}{1.4cm}{\textit{Retriever Adaptation}} & \textit{Disjoint} & \textit{End-to-End} & \textit{Multi-Step} & \multirow{2}{1.7cm}{\textit{Unsupervised Retriever}} \\
 & & & & & & \\
 \midrule
 REALM~\citep{guu2020realm} &  & \cmark  &  &  \cmark &  & \cmark \\
 DPR~\citep{karpukhin2020dense} &  & & \cmark &  &  \\
 RAG~\citep{lewis20rag} &  & \cmark &  & \cmark &  \\
 FiD~\citep{izacard2021leveraging} & \cmark & & \cmark &  &  \\
 FiD-KD~\citep{izacard2021distilling} & \cmark & \cmark &  &  & \cmark \\
 \midrule
 \emdr{} (Our Approach) & \cmark & \cmark &  & \cmark &  & \cmark \\
 \bottomrule
\end{tabular}
\caption{Bird's-eye view of the recent OpenQA approaches. \textbf{Multi-Doc reader} indicates whether the reader architecture uses multiple documents or a single document. \textbf{Retriever adaptation} shows whether the retriever gets feedback from the reader to update its parameters. \textbf{Disjoint} denotes that first the retriever is trained and then the reader is trained.
\textbf{End-to-end} denotes that the reader and retriever are trained jointly in one cycle. \textbf{Multi-step} indicates that the reader and retriever are trained iteratively in multiple cycles.
\textbf{Unsupervised retriever} indicates whether the retriever is initialized using unsupervised approaches or using supervised data. 
}
\label{tab:cmp}
\end{table*}


\section{Model}
\label{sec:model}
\begin{figure*}[t]
\centering
\includegraphics[max width=1.0\textwidth, scale=0.9,trim=0cm 0cm 2.5cm 0cm, clip=true]{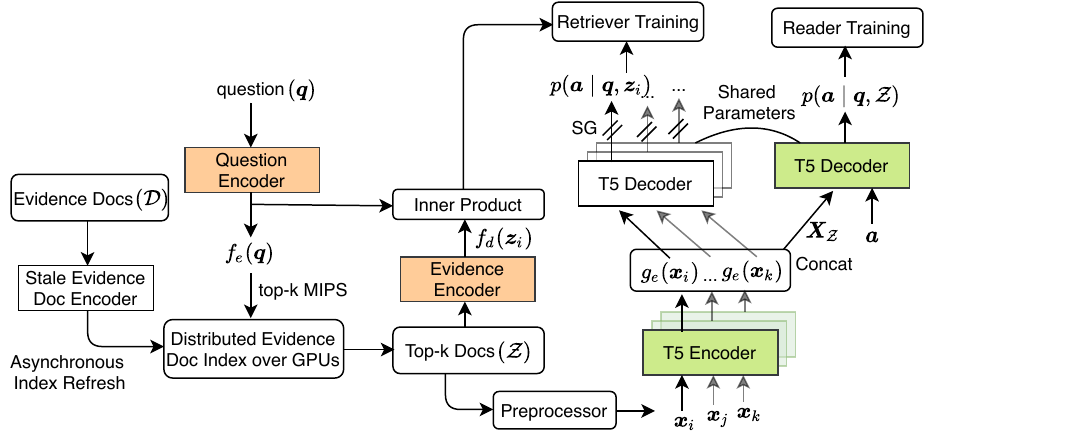} 
\caption{An illustration of the different components of \emdr{}. Colored blocks indicate components which contain trainable parameters.}
\label{fig:end-to-end-training}
\end{figure*}

Our proposed model \emdr{} consists of two components: (i) a neural retriever and (ii) a neural reader, which we train jointly in an end-to-end setting. Figure~\ref{fig:end-to-end-training} shows an illustration of our model and training procedure. We discuss each component and our training objective in detail below.

\subsection{Neural Retriever: Dual Encoder}
Let the collection of evidence documents be denoted by $\set{D} = \{\boldsymbol{d}_1, \ldots, \boldsymbol{d}_M \}$.
Given a question $\boldsymbol{q}$, the goal of the retriever module is to select a subset of documents $\set{Z} \subset \set{D}$ to answer the question.
We model the retriever as a dual-encoder network~\citep{bromley1994signature}, where one encoder $f_q$ encodes the question and another $f_d$ encodes the evidence document (to a vector). The retrieval score is defined as the dot product between the two resulting vectors:
\begin{align}
\label{eq:score}
\text{score}(\boldsymbol{q}, \boldsymbol{d}_i; \Phi) = f_{q}(\boldsymbol{q}; \Phi_q)^\top f_{d}(\boldsymbol{d}_i; \Phi_d),
\end{align}
where $\Phi=[\Phi_q, \Phi_d]$ denotes the retriever parameters.
We select top-$K$ documents for the question $\boldsymbol{q}$ from $\mathcal{D}$ based on the retrieval scores. We denote the set of retrieved documents by $\mathcal{Z} = \{\boldsymbol{z}_1, \ldots, \boldsymbol{z}_K\}$.

We use transformer encoders~\citep{vaswani2017attention} as our
$f_{q}$ and $f_{d}$. Our transformer architecture is similar to BERT with 12 layers and 768 hidden size \citep{devlin2019bert}. We use the final representation of the first token (i.e., the standard \texttt{[CLS]} token from BERT's tokenization) as our 
question (and similarly document) embedding. 
Initializing $f_{q}$ and $f_{d}$ with BERT weights has been shown to lead to a poor retrieval accuracy~\citep{lee-etal-2019-latent, sachan2021end}. 
Therefore, we initialize the retriever with an unsupervised training procedure. We discuss our initialization technique in detail in \S\ref{sec:impl-details}.

\subsection{Neural Reader: Fusion-in-Decoder}

The reader takes as input a question $\boldsymbol{q}$ and a set of retrieved documents (to be read) $\mathcal{Z}$ to generate an answer.
Our reader is based on the Fusion-in-Decoder (FiD; \citealp{izacard2021leveraging}) model,
which is built on top of T5~\citep{raffel2020t5}. 
T5 is a pretrained sequence-to-sequence transformer that consists of an encoder $g_{e}$ and a decoder $g_{d}$.

In FiD, each retrieved document $\boldsymbol{z}_k$ is first 
appended with its title ($\boldsymbol{t}_{\boldsymbol{z}_k}$) and the question: 
\begin{align*}
\boldsymbol{x}_k = \texttt{[CLS]} \boldsymbol{q} \texttt{[SEP]} \boldsymbol{t}_{\boldsymbol{z}_k} \texttt{[SEP]} \boldsymbol{z}_k \texttt{[SEP]},
\end{align*}
where \texttt{[CLS]} is used to indicate the start of a document and
\texttt{[SEP]} is used as a separator for the different parts of the document
as well as the final token.

Each $\boldsymbol{x}_k$ is then independently given as an input to the T5 encoder $g_e$. The output representations corresponding to all of the retrieved documents are concatenated as:
\begin{align*}
\vect{X}_{\mathcal{Z}} = [g_{e}(\boldsymbol{x}_1);\ldots; g_{e}(\boldsymbol{x}_K)] \in \mathbb{R}^{(N \times K) \times H},
\end{align*}
where $N$ is the number of tokens in each $\boldsymbol{x}_k$\footnote{We truncate and pad as necessary such that every $\boldsymbol{x}_k$ has the same length $N$. See \S\ref{sec:impl-details} for details.}
and $H$ is the hidden size of the T5 encoder $g_e$. 
In this work, we use the T5-\emph{base} configuration with $N=512$ and $H=768$.

$\vect{X}_{\mathcal{Z}}$ is then given as an input to the T5 decoder $g_d$. When generating an answer token, 
the decoder attends to both previously generated tokens (i.e., causal attention) as well as the tokens encoded in $\vect{X}_{\mathcal{Z}}$ (i.e., cross attention).
Since $\vect{X}_{\mathcal{Z}}$ contains information from multiple documents,
the decoder has the ability to aggregate useful signals contained in 
multiple documents and jointly reason over them.
We define the probability of the answer as:
\begin{align}
\label{eq:generateanswer}
p(\boldsymbol{a} \mid \boldsymbol{q}, \mathcal{Z}; \Theta) = \prod_{t=1}^T p\left(a_t \mid \boldsymbol{a}_{<t}, \boldsymbol{q}, \mathcal{Z};\Theta\right),
\end{align}
where $\Theta$ denotes the reader parameters (i.e., T5 encoder and decoder)
and $T$ is the number of answer tokens.
We keep generating answer tokens until the decoder outputs a special \texttt{EOS} token
or a pre-specified maximum answer length is reached.

\subsection{End-to-End Training of Reader and Retriever}
\label{sec:trainingobj}

In contrast to previous work on generative question answering, we train
both the reader and the retriever jointly in an end-to-end differentiable fashion.

Denote our latent variable which represents a set of retrieved documents by $Z$ 
and let $\mathcal{Z}$ be a possible value of $Z$.
The marginal likelihood of an answer
(marginalizing over all the possible values of $Z$)
is:
$p(\boldsymbol{a} \mid \boldsymbol{q}; \Theta, \Phi) = \sum_{Z = \mathcal{Z}} p(\boldsymbol{a} \mid \boldsymbol{q}, \mathcal{Z}; \Theta)p(\mathcal{Z} \mid \boldsymbol{q}; \Phi)$.
The goal of our training procedure is to find $\Phi$ and $\Theta$ that would maximize the above objective.
Exactly optimizing Eq.~\ref{eq:obj} is intractable as it is combinatorial in nature.\footnote{
Contrast our objective with REALM \citep{guu2020realm}, where the reader only conditions on one 
retrieved document $\boldsymbol{z}_k$ when generating an answer. In this case, the latent variable represents a document assignment instead of a set of retrieved documents.}
For one particular value $\mathcal{Z}$, the log-likelihood is simpler to compute:
$\log p(\boldsymbol{a} \mid \boldsymbol{q}, \mathcal{Z}; \Theta) p(\mathcal{Z} \mid \boldsymbol{q}; \Phi) = 
\log p(\boldsymbol{a} \mid \boldsymbol{q}, \mathcal{Z}; \Theta) + \log p(\mathcal{Z} \mid \boldsymbol{q}; \Phi)$.

Expectation-maximization (EM) algorithm \citep{em} offers a solution to learning this
latent variable model. In classical EM, we iteratively compute the posterior of $Z$ given all
observed variables and use it to update $\Theta$ and $\Phi$.

We propose using two estimates of $Z$---$\mathcal{Z}_{\text{reader}}$ and $\mathcal{Z}_{\text{retriever}}$---for updating the two components of the model (reader parameters $\Theta$ and retriever parameters $\Phi$):
\begin{align}
\label{eq:obj}
\log \underbrace{p(\boldsymbol{a} \mid \boldsymbol{q}, \mathcal{Z}_{\text{reader}}; \Theta)}_\textrm{reader} + \log \underbrace{p(\mathcal{Z}_{\text{retriever}} \mid \boldsymbol{q}; \Phi)}_{\textrm{retriever}}.
\end{align}
In the first term, we set the value of the latent variable $Z = \mathcal{Z}_{\text{reader}}$
based on the prior scores. In the second term, we seek to maximize an approximate posterior of $Z = \mathcal{Z}_{\text{retriever}}$.
We discuss them in more detail below.

\paragraph{Reader parameters $\Theta$.}
For updating $\Theta$ (the first term of Eq.~\ref{eq:obj}), we use the top-$K$ documents with the highest individual
scores (as computed by Eq.~\ref{eq:score} based on 
the current value of $\Phi$) to construct $\mathcal{Z}_{\text{reader}}$. This is equivalent to relying on
the prior $p(Z \mid \boldsymbol{q}; \Phi)$ to estimate $\mathcal{Z}_{\text{reader}}$ (without using 
information from the answer $\boldsymbol{a}$).
We choose to use the prior to train reader parameters since the prior scores are also used at evaluation time to obtain the top-$K$ documents.
As a result, there is no mismatch between training and test computations
when computing $p(\boldsymbol{a} \mid \boldsymbol{q}, \mathcal{Z}; \Theta)$ (i.e., $\mathcal{Z}$
that is used at test time is obtained in exactly the same way as $\mathcal{Z}_{\text{reader}} = \mathcal{Z}_{\text{top-}K}$).

\paragraph{Retriever parameters $\Phi$.}
For updating $\Phi$ (the second term of Eq.~\ref{eq:obj}), we propose to use
the posterior estimate. 
In other words, we use additional information from $\boldsymbol{a}$ 
when evaluating $Z_{\text{retriever}}$ to train $\Phi$.
Using the posterior allows our retriever to learn from richer training signals
as opposed to relying only on the prior. 

We need to be able to compute 
$p(\mathcal{Z}_{\text{retriever}} \mid \boldsymbol{q},\boldsymbol{a}; \Theta, \Phi)$
to maximize the retriever parameters. However, computing this quantity is difficult since it is a probability of a set.\footnote{This is true whether we choose to use the posterior probability or the prior probability.}
Consider a set of $K$ documents (e.g., $\mathcal{Z}_{\text{top-}K}$), where $\boldsymbol{z}_k$ denotes a document in the set.
We approximate the maximization of the probability of the set by assuming that its probability is maximized if the sum of the probability of each document in the set is maximized.\footnote{The intuition is that each element of the set contributes independently, which greatly simplifies the computation to find the maximum of the set.} 
With this approximation, we arrive at a simpler quantity: $\sum_{k=1}^K p(\boldsymbol{z}_k \mid \boldsymbol{q},\boldsymbol{a}; \Theta, \Phi)$.
Note that using Bayes rule, we can rewrite:\footnote{We choose not to normalize with $p(\boldsymbol{a} \mid \boldsymbol{q}; \Theta, \Phi)$ since computing this quantity would require summing over all evidence documents $M$. While this makes the resulting objective that we optimize not correspond to a proper probability distribution anymore, we observe that our training method still behaves well in practice.
}
\begin{align}
\label{eq:bayes}
p(\boldsymbol{z}_k \mid \boldsymbol{q},\boldsymbol{a}; \Theta, \Phi) \propto p(\boldsymbol{a} \mid \boldsymbol{q},\boldsymbol{z}_k; \Theta)p(\boldsymbol{z}_k\mid \boldsymbol{q}; \Phi).
\end{align}
The reader now only conditions on one document when computing
the probability of an answer $p(\boldsymbol{a}\mid\boldsymbol{q}, \boldsymbol{z}_k; \Theta)$.
This simpler reader uses the same parameters as the more sophisticated one $\Theta$, but it only uses one document $\boldsymbol{z}_k$ instead of a set of documents.

To compute Eq.~\ref{eq:bayes}, we first 
obtain $K$ documents with the highest scores as computed by Eq.~\ref{eq:score} based on the current value of $\Phi$.
We compute the probability of document $\boldsymbol{z}_k \in \mathcal{Z}_{\text{top-}K}$ as:
\begin{align}
\label{eq:ret-prob}
p(\boldsymbol{z}_k \mid \boldsymbol{q}, \mathcal{Z}_{\text{top-}K}; \Phi) \approx \frac{\exp (\text{score}(\boldsymbol{q}, \boldsymbol{z}_k)/\tau; \Phi)}{\sum_{j=1}^K \exp (\text{score}(\boldsymbol{q}, \boldsymbol{z}_j)/\tau; \Phi)},
\end{align}
where $\tau$ is a temperature hyperparameter and the approximation 
assumes that documents beyond the 
top-$K$ contributes very small scores so we 
do not need to sum over all evidence documents 
$M$ in the denominator (which is in the order of tens of millions in our experiments).
We then compute $p(\boldsymbol{a}\mid\boldsymbol{q}, \boldsymbol{z}_k; \Theta)$ similarly to Eq.~\ref{eq:generateanswer}.

\paragraph{Overall training objective of \emdr{}.}
Combining the above derivations, our end-to-end training 
objective that we seek to maximize for a particular example becomes:
\begin{align}
\label{eq:realobj}
\mathcal{L} = \underbrace{\log p(\boldsymbol{a} \mid \boldsymbol{q}, \mathcal{Z}_{\text{top-}K}; \Theta)}_\textrm{reader} + \underbrace{\log \sum_{k=1}^{K} \mathbb{SG}\left({p}(\boldsymbol{a} \mid \boldsymbol{q}, \boldsymbol{z}_{k}; \Theta)\right) p(\boldsymbol{z}_k \mid \boldsymbol{q}, \mathcal{Z}_{\text{top-}K}; \Phi)}_\textrm{retriever},
\end{align}
where $\mathbb{SG}$ is the stop-gradient operator so that the reader parameters $\Theta$ are not updated to also perform well given a single document $\boldsymbol{z}_k$.
The stop-gradient operator in the second term of \emdr{} has several benefits.
First, the FiD reader is trained from the first term of the \emdr{} objective in which its likelihood is conditioned on all the retrieved documents, similar to how the reader is used at test time. 
Second, it also makes training faster since the backward pass which is computationally more expensive than the forward pass is not needed, which in turn reduces the usage of GPU RAM as intermediate activations need not be saved.

Given a training example, we update $\Theta$ and $\Phi$ by taking gradients of Eq.~\ref{eq:realobj} with respect to $\Theta$ and $\Phi$ in an end-to-end fashion.
Intuitively, we train the reader to generate the correct answer given $K$ highest scoring documents $\mathcal{Z}_{\text{top-}K}$. For the retriever, we train it to select $K$ documents which
\emph{collectively} has a high score of generating an answer (since the sum over $K$ is inside the log in the second term) while taking into account feedback from the reader. Algorithm~\ref{tab:em-algorithm} summarizes our training algorithm.

\begin{algorithm}[h]
\SetAlgoLined
\DontPrintSemicolon
\KwIn{Model parameters $\Theta$ and $\Phi$, evidence documents $\mathcal{D}$.}
\While{not converged}{
\tabitem Compute $\mathcal{Z}_{\text{top-}K}$ using the current retriever parameters $\Phi$. \tcp*{E-step}
\tabitem Compute ${p}(\boldsymbol{a} \mid \boldsymbol{q}, \boldsymbol{z}_{k})$ for each $\boldsymbol{z}_{k}$ using the current reader parameters $\Theta$. \tcp*{E-step}
\tabitem Update model parameters $\Theta$ and $\Phi$ to maximize the log-likelihood in Eq.~\ref{eq:realobj}. 
\tcp*{M-step}
}
 \caption{End-to-end training of multi-document reader and retriever.}
 \label{tab:em-algorithm}
\end{algorithm}


\section{Experiments}
\label{sec:exp-setup}

\subsection{Datasets}
\label{sec:dataset}
We experiment with three commonly used open-domain question answering datasets:
\begin{itemize}
\item \textbf{Natural Questions (NQ; \citealp{Kwiatkowski2019natural}).} NQ contains questions asked by users of the Google search engine. 
Similar to \citet{lee-etal-2019-latent}, we use the short answer subset.
\item \textbf{TriviaQA \citep{joshi2017triviaqa}.} TriviaQA is a collection of trivia question-answer pairs that were collected from multiple sources on the web.
\item \textbf{WebQuestions (WebQ; \citealp{berant-etal-2013-semantic}).} WebQ questions were collected using Google Suggest API and the answers were annotated using Mechanical Turk. We use the version from \citet{chen-etal-2017-reading} where Freebase IDs in the answers are replaced by entity names.
\end{itemize}

\paragraph{Evidence documents $\mathcal{D}$.} We use the preprocessed English Wikipedia dump from December 2018 released by \citet{karpukhin2020dense} as our evidence documents. Each Wikipedia article is split into non-overlapping 100 words long segments. Each segment corresponds to a document in our case. There are a total of 21,015,324 documents in total.

We provide descriptive statistics and other preprocessing details in Appendix~\ref{appendix:dataset-stat}.

\subsection{Implementation Details}
\label{sec:impl-details}
\paragraph{Hardware and library.}
We run all of our experiments on a machine with 96 CPUs, 1.3TB physical memory, and 16 A100 GPUs.
We use PyTorch \citep{paszke2019pytorch} to implement our proposed model and relevant baselines.

\paragraph{Model configurations.}
For both the retriever and reader, we use the \emph{base} configuration that consists of 12 layers, 768 dimensional hidden size, and 12 attention heads. In all experiments, we retrieve 50 documents, unless stated otherwise. We only use the base configuration in our experiments due to GPU memory constraints. However, we believe that our results would generalize to larger configurations as well.

\paragraph{Retrieval.}
To support fast retrieval, we pre-compute evidence document embeddings and store them in a distributed fashion over all the GPUs. We refer to these document embeddings as the document index.
For each question, we retrieve documents in an online (on-the-fly) manner by performing exact maximum inner product search (MIPS), implemented using asynchronous distributed matrix multiplication over the document index.
These documents are converted to subwords using BERT's tokenization and are given as input to the T5 reader. If a tokenized document is shorter than 512 tokens, it is padded using the tokens from the neighboring documents until the maximum token limit is reached. Such padding additionally helps to provide an extended context for answer generation.

\paragraph{Initialization and training details.}
We initialize the parameters of the model with unsupervised pre-training before performing supervised training using the question-answer training examples. 
Unsupervised pre-training is essential as it helps to warm-start 
the retriever so that it outputs relevant documents for a given question.

We first pre-train the retriever parameters with unsupervised Inverse Cloze Task training~\citep{lee-etal-2019-latent} for 100,000 steps.
We then extract sentences containing named entities from the evidence documents. 
Next, we replace 15\% of the named entity tokens with masked tokens, which are often referred to as masked salient spans (MSS; \citealp{guu2020realm}).
The masked sentence can be considered as the question and its salient spans (i.e, named entities) can be considered as the answer to train the model with Eq.~\ref{eq:realobj}.
We train the model on these question-answer (masked sentence-named entities) pairs for 82,000 steps with a batch size of 64 using Adam~\citep{kingma2014adam}.
We refer to this initialization method as \emph{unsupervised pre-training with masked salient spans}. We provide further description in Appendix~\ref{sec:realm-cmp}.

After MSS training, we finetune the model on the 
dataset-specific question-answer training examples with \emdr{}. 
We perform training for 10 epochs on NQ and TriviaQA with a 
batch size of 64, and for 20 epochs on WebQ with a batch size of 16. During training, we save a checkpoint every 500 steps and select the best checkpoint based on its performance on the development set.

During end-to-end training, since the parameters of the document encoder ($f_d$) are also updated at every step, the pre-computed document embeddings become stale as training progresses. 
We use the most recent document encoder checkpoint to compute fresh document embeddings asynchronously with which the document index is updated after every 500 training steps to prevent
staleness.

\paragraph{Inference.} We use greedy decoding for answer generation at inference time.


\subsection{Baselines}
We compare our model to other approaches for OpenQA that can 
be categorized under the following two classes:

\begin{itemize}
\item \textbf{Closed-book QA models.} Large-scale language models capture a lot of world knowledge in their parameters derived from the corpus they have been trained on~\citep{petroni-etal-2019-language}. We compare with the work of~\cite{roberts-etal-2020-much} who show that larger T5 models---when finetuned with question-answer pairs---can perform remarkably well. We also compare with the few-shot results of GPT-3 \citep{brown2020gpt3}.\footnote{We note that GPT-3 is not trained on the full training examples that we use, so the results are not directly comparable.}

\item \textbf{Open-book QA models.} Similar to this work, these models consist of retriever and reader components and adopt the retrieve then predict approach for answering questions given a collection of evidence documents. 
These models mainly differ in how the retriever is initialized (ORQA;~\citealp{lee-etal-2019-latent}, DPR;~\citealp{karpukhin2020dense}), whether the reader processes a single document (ORQA, DPR, RAG;~\citealp{lewis20rag}) or multiple documents (FiD; \citealp{izacard2021leveraging}), or whether the reader and retriever are trained jointly or in a multistage process (REALM;~\citealp{guu2020realm}, FiD-KD;~\citealp{izacard2021distilling}).
\end{itemize}

\subsection{Results}
\label{sec:results}

\begin{table*}[t]
\small
\centering
\begin{tabular}{l c c c c c c c c}
 \toprule
 \textbf{Model} & top-$K$ & \multicolumn{2}{c}{\textbf{NQ}} & \multicolumn{2}{c}{\textbf{TriviaQA}} & \multicolumn{2}{c}{\textbf{WebQ}} & \textit{\# of} \\
                &    &  dev & test & dev & test & dev & test & \textit{params}\\
 \midrule
 \multicolumn{9}{c}{\textbf{Closed-Book QA Models}} \\
 \midrule
 T5-\emph{base}~\citep{roberts-etal-2020-much} & \phantom{00}0 & - & 25.7 & - & 24.2 & - & 28.2 & 220M \\
 T5-\emph{large}~\citep{roberts-etal-2020-much} & \phantom{00}0 & - & 27.3 & - & 28.5 & - & 29.5 & 770M \\
 T5-\emph{XXL}~\citep{roberts-etal-2020-much} & \phantom{00}0 & - & 32.8 & - & 42.9 & - & 35.6 & \phantom{0}11B \\
 GPT-3~\citep{brown2020gpt3} & \phantom{00}0 & - & 29.9 & - & - & - & 41.5 & 175B \\
 \midrule
 \multicolumn{9}{c}{\textbf{Open-Book QA Models}} \\
\midrule
 BM25 + BERT~\citep{lee-etal-2019-latent}& \phantom{00}5& 24.8 & 26.5 & 47.2 & 47.1 & 27.1 & 21.3 & 220M \\
 ORQA~\citep{lee-etal-2019-latent}     & \phantom{00}5 & 31.3 & 33.3 & 45.1 & 45.0 & 36.8 & 30.1 & 330M \\
 REALM~\citep{guu2020realm}            & \phantom{00}5 & 38.2 & 40.4 &  -   &  -   &  -   & 40.7 & 330M \\
 DPR~\citep{karpukhin2020dense}        & \phantom{0}25 &  -   & 41.5 &  -   & 56.8 &  -   & 34.6 & 330M \\
 \textsc{ReConsider}~\citep{iyer2020reconsider}$\dagger$ &  \phantom{0}30 &  -   & 43.1 &  -   & 59.3 & -    & 44.4 & 440M \\
 RAG-Sequence~\citep{lewis20rag}$\dagger$       & \phantom{0}50 & 44.0 & 44.5 & 55.8 & 56.8 & 44.9 & 45.2 & 626M \\
 Individual Top-$K$~\citep{sachan2021end}&      -        &  -   & 45.9 &  -   & 56.3 &  -   &  -   & 440M \\
 Joint Top-$K$~\citep{sachan2021end}     & \phantom{0}50 &  -   & 49.2 &  -   & 64.8 &  - & - & 440M \\
 FiD~\citep{izacard2021leveraging}     & 100           &  -   & 48.2 &  -   & 65.0 & -  & - & 440M \\
 FiD-KD~\citep{izacard2021distilling}  & 100           & 48.0 & 49.6 & 68.6 & 68.8 &  -  & - & 440M \\
\midrule
\multicolumn{9}{c}{\textbf{Our Implementation (Base Configuration)}} \\
\midrule
FiD / T5-\emph{base}    & \phantom{00}0 & 26.0 & 25.1 & 26.7 & 27.8 & 31.0 & 32.4 & 220M \\
FiD (DPR retriever, T5 reader)               & \phantom{00}1 & 37.3 & 38.4 & 50.8 & 50.4 & 40.2 & 38.3 & 440M \\
FiD (DPR retriever, T5 reader)               & \phantom{0}50 & 47.3 & 48.3 &  65.5 & 66.3 & 46.0 & 45.2 & 440M \\
FiD (MSS + DPR retriever, T5 reader)         & \phantom{0}50 & 48.8 & 50.4 & 68.0 & 68.8 & 43.5 & 46.8 & 440M \\
\midrule
FiD (MSS retriever, MSS reader)            & \phantom{0}50 & 38.5 & 40.1 & 60.0 & 59.8 & 39.1 & 40.2 & 440M \\
$\text{EMDR}^2$ (MSS retriever, MSS reader) & \phantom{0}50 & \textbf{50.4} & \textbf{52.5} & \textbf{71.1} & \textbf{71.4} & \textbf{49.9} & \textbf{48.7} & 440M \\
 \bottomrule
\end{tabular}
\caption{Exact match scores on three evaluation datasets. Top-$K$ denotes the number of retrieved documents that are used by the reader to produce an answer.
To provide a fair comparison with our reimplementations, we show results from other papers with the base configuration, except for RAG-Sequence that uses BART-\emph{large}~\citep{lewis2020bart}.
$\dagger$ indicates that their results on WebQ use NQ training data to pretrain the model.
}
\label{tab:answer-extraction-results}
\end{table*}

We follow standard conventions and report exact match (EM) scores using the reference answers included in each dataset. 
Table~\ref{tab:answer-extraction-results} shows our main results. We divide the table into
three main sections: closed-book QA models, open-book QA models, and our implementation. The
first two sections contain results from other papers, which we include for comparisons.
The last section includes results from our proposed model, as well as our reimplementation
of relevant baselines to control for our experimental setup.

Our reimplementation of T5-base provides strong baselines when the number of retrieved documents is set to 0 (no retrieval) and 1.
From Table~\ref{tab:answer-extraction-results}, we see that the setting of top-$1$ vastly improves performance over the setting with no retrieved documents, signifying the importance of retrieval for OpenQA tasks. When further increasing the top-$k$ documents to 50, the performance of the FiD models substantially improves over the top-$1$ retrieval, verifying the observation from~\citep{izacard2021leveraging} about the \emph{importance of modeling the retrieved documents as a set}.

Comparing \emdr{} with our reimplementation of FiD illustrates 
the benefit of our end-to-end training approach.
The underlying model is similar in both cases, but the training method is different.
FiD adopts a two-stage approach to first train the retriever and then the reader.
We have three variants of FiD: (i) the reader and retriever are initialized with MSS training, (ii) the retriever is initialized with DPR training, which is the setting used in the original paper~\citep{izacard2021leveraging}, and (iii) the retriever is initialized with MSS + DPR training from~\citep{sachan2021end}, as it further improves DPR recall. 
\emdr{} outperforms all the variants by large margins on all the datasets. 

The current best approach for training multi-document reader and retriever is FiD-KD \citep{izacard2021distilling}. FiD-KD is a complex training procedure that requires multiple training stages and performs knowledge distillation with inter-attention scores.
We take the results from the original paper when comparing our model with FiD-KD.
\emdr{} outperforms the reported numbers of FiD-KD by more than 2.5 points on NQ and TriviaQA to obtain new state-of-the-art results on these benchmarks.

In addition to better performance, \emdr{} also has three other advantages compared to FiD-KD:
(i) \emdr{} is more efficient since it only uses 50 evidence documents, whereas FiD-KD leverages 100 documents; (ii) FiD-KD is based on a distillation approach which requires multiple cycles of retriever and reader training, while \emdr{} only requires one cycle of end-to-end training;
and (iii) FiD-KD relies on supervised initialization of the retriever to achieve its best performance. \emdr{} is more robust to the retriever initialization, as demonstrated by state-of-the-art results even with unsupervised initialization of the retriever.

For the WebQ dataset, the training set size is much smaller compared to the other datasets (Table~\ref{table:dataset-stat}). Previous approaches such as RAG rely on supervised transfer (i.e., they finetune a model pre-trained on NQ) to obtain good results. In contrast, \emdr{} improves over the results from this RAG model by 3.5 points \emph{without the supervised transfer step}.
This result demonstrates the applicability of our approach to the low-resource setting where we only have a limited number of training examples.

We also perform qualitative analysis of the model outputs, which is included in Appendix~\ref{sec:qual-analysis}.
\subsection{Ablations}
\label{sec:ablation}

\begin{figure*}[tb]
\centering
\includegraphics[max width=0.33\textwidth, scale=0.99]{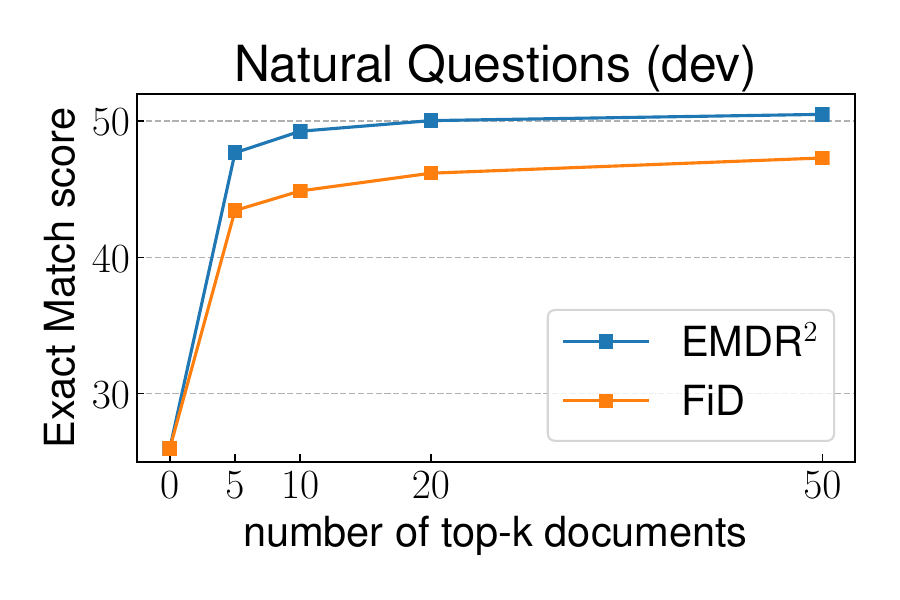}
\includegraphics[max width=0.33\textwidth, scale=0.99]{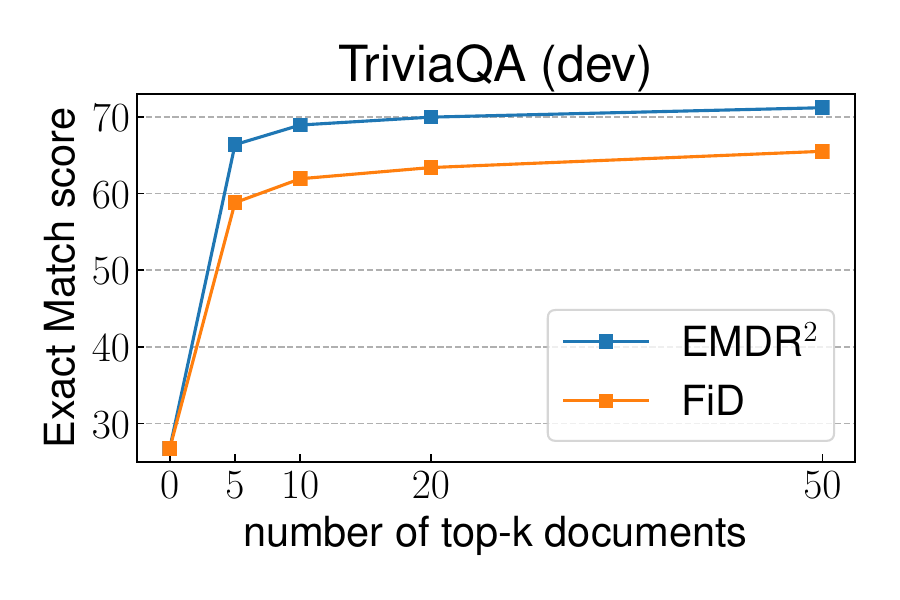}
\includegraphics[max width=0.33\textwidth, scale=0.99]{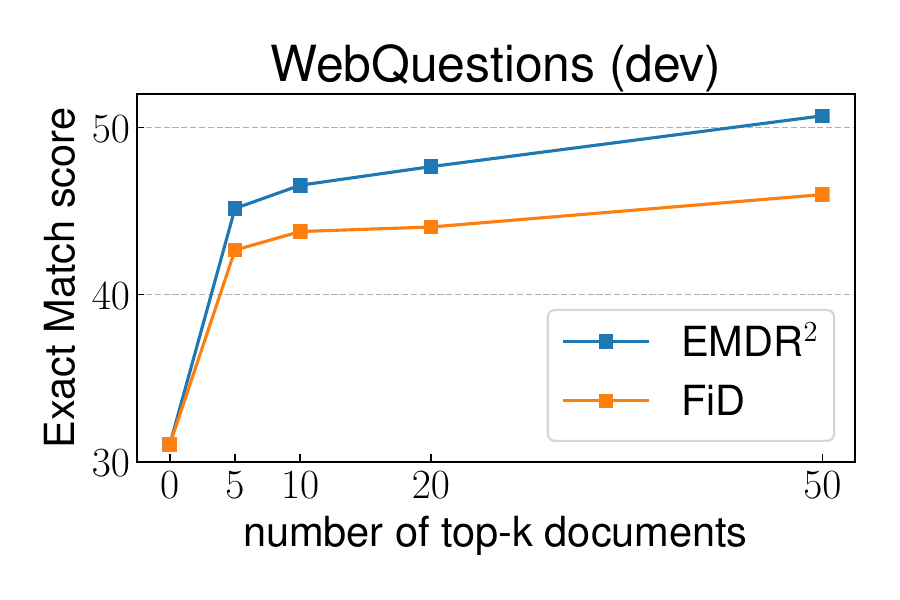}
\caption{Performance on NQ, TriviaQA, and WebQ as we vary the number of retrieved documents.}
\label{fig:topk-em}
\end{figure*}

\paragraph{Number of retrieved documents.}
We investigate the performance of \emdr{} and FiD as we vary the number of retrieved documents $K$ in Figure~\ref{fig:topk-em}. We observe that when the number of retrieved documents is increased, both \emdr{} and FiD improve in performance. When $K$ is small, the gap between \emdr{} and FiD
is larger. This indicates the efficacy of \emdr{} in a more constrained setting
where we can only retrieve a small number of documents (e.g., due to memory limitations).

\textbf{Retriever initialization.} 
We explore the effect of different parameter initialization strategies when training with \emdr{}: (i) unsupervised MSS pre-training, (ii) supervised retriever training (DPR), and (iii) MSS pre-training followed by supervised retriever training (MSS + DPR; \cite{sachan2021end}).
Table~\ref{tab:ret-init} shows our results. 
We can see that on NQ, MSS pre-training being unsupervised leads to a lower initial retriever recall than DPR. 
After \emdr{} training, the recall improves by 20\% (highlighted in yellow cells). 
Training with DPR initialization leads to the same final recall as obtained by MSS pre-training, \emph{suggesting that DPR initialization of the retriever may not be an essential component to obtain good performance in OpenQA tasks.} 
Similar trends are also observed on TriviaQA and WebQ. 
Similarly, MSS + DPR initialization has a better initial recall but leads to a marginal or no improvements in answer extraction performance over MSS pre-training. 
Finally, we also observe that MSS pre-training also provides an improvement of 2 points in answer extraction on WebQ when compared to the T5 reader (shown in orange cells), highlighting its importance in the low-resource OpenQA tasks.

\begin{table*}[t]
\small
\centering
\begin{tabular}{l l c c c c c c c c c c}
\toprule
&  & \multicolumn{3}{c}{\textbf{NQ (dev)}} & \multicolumn{3}{c}{\textbf{TriviaQA (dev)}} & \multicolumn{3}{c}{\textbf{WebQ (dev)}} \\
\midrule
\multirow{2}{2cm}{\textbf{Retriever Initialization}} & \multirow{2}{2cm}{\textbf{Reader Initialization}} & \multicolumn{2}{c}{\textbf{R@50}} & \textbf{EM} & \multicolumn{2}{c}{\textbf{R@50}} & \textbf{EM} & \multicolumn{2}{c}{\textbf{R@50}} & \textbf{EM} \\
  & & B.T. & A.T. &  & B.T. & A.T. &  & B.T. & A.T. & \\
\midrule
MSS pre-training & MSS pre-training & {\cellcolor{Yellow}66.4} & {\cellcolor{Yellow}86.3} & 50.4 & 74.8 & 86.2 & 71.1 & 59.8 & 88.6 & {\cellcolor{Orange}49.9}   \\
MSS pre-training & T5               & 66.4 & 86.3 & 50.3 & 74.8 & 86.3 & 70.9 & 59.8 & 88.6 & {\cellcolor{Orange}47.7} \\
DPR training     & T5               & 82.3 & 86.3 & 50.0 & 83.2 & 86.2 & 70.5 & 84.2 & 88.6 & 49.0 \\
MSS + DPR        & MSS pre-training & 84.5 & 86.3 & 50.5 & 85.3 & 86.3 & 71.2 & 85.0 & 88.6 & 49.9 \\
\bottomrule
\end{tabular}
\caption{R@50 denotes the retrieval recall from the top-$50$ retrieved documents. B.T. and A.T. indicates R@50 score Before Training and After Training the model, respectively.}
\label{tab:ret-init}
\end{table*}

\subsection{Alternative End-to-End Training Objectives}
We compare \emdr{} objective (Eq.~\ref{eq:realobj}) to two alternative formulations for end-to-end training. 

\begin{wraptable}{r}{0.4\textwidth}
\vspace{0mm}
\small
\centering
\setlength{\tabcolsep}{0.30em}
\begin{tabular}{l c c c c}
\toprule
\textbf{Method} & top-$k$ & \textbf{NQ} & \textbf{TriviaQA} & \textbf{WebQ} \\
\midrule
FiD & 50 & 47.3 & 65.5 & 46.0 \\
\emdr{} & 50 & \textbf{50.4} & \textbf{71.1} & \textbf{49.9} \\
\midrule
$\mathcal{L}_{\text{alt-1}}$ & 50 & 14.1 & 11.9 & 28.0 \\
$\mathcal{L}_{\text{alt-2}}$ & 50 & 49.9 & 69.6 & 28.8 \\
\bottomrule
\end{tabular}
\caption{EM scores on the development set for alternative training objectives.}
\label{tab:obj-func}
\vspace{-5mm}
\end{wraptable}

In the first alternative formulation, when training the retriever parameters $\Phi$,
we simply factorize $p(\mathcal{Z} \mid \boldsymbol{q}; \Phi) = \prod_{k=1}^K p(\boldsymbol{z}_k \mid \boldsymbol{q}; \Phi)$ to arrive at the following objective:
\begin{align*}
\mathcal{L}_{\text{alt-1}} = \log p(\boldsymbol{a} \mid \boldsymbol{q}, \mathcal{Z}; \Theta) + \sum_{k=1}^{K} \log p(\boldsymbol{z}_k \mid \boldsymbol{q}, \mathcal{Z}; \Phi).
\end{align*}
The second term in this objective is maximised by a uniform retrieval, in other words, by \emph{removing} any discrimination between documents in the retriever. We include it to show the impact of an adversarial objective.

In the second formulation, for each retrieved document, we approximate its posterior 
under the assumption that we have a uniform prior over the set of retrieved documents:
$\tilde{p}(\boldsymbol{z}_k \mid \boldsymbol{q}, \boldsymbol{a}, \mathcal{Z}_{\text{top-$K$}}; \Theta) \propto p(\boldsymbol{a} \mid \boldsymbol{q}, \boldsymbol{z}_{k}; \Theta) \times \frac{1}{K}$.
We use this to train reader and retriever parameters as follows:
\begin{align*}
\mathcal{L}_\text{alt-2} = \log p(\boldsymbol{a} \mid \boldsymbol{q}, \mathcal{Z}; \Theta) + \kld{\mathbb{SG}\left(\tilde{p}(\boldsymbol{z}_k \mid \boldsymbol{q}, \boldsymbol{a}, \mathcal{Z}_{\text{top-$K$}}; \Theta)\right)}{p(\boldsymbol{z}_k \mid \boldsymbol{q}, \mathcal{Z}; \Phi)}.
\end{align*}
Intuitively, we try to match the probability of retrieving a document $\boldsymbol{z}_k$
with the ``contribution'' of that document to the generated answer $\boldsymbol{a}$, regardless of whether the retriever is relatively more or less likely to retrieve the document \emph{a priori}.

Table~\ref{tab:obj-func} shows our results on the development set of NQ. We observe 
that training with the adversarial $\mathcal{L}_{\text{alt-1}}$ objective diverges, leading to poor performance, as expected. This shows that harming the retriever during training can significantly harm performance of the QA system.
In contrast, although it disregards the estimated prior, the $\mathcal{L}_{\text{alt-2}}$ objective still improves over the FiD baseline for NQ and TriviaQA. However, it still lags behind \emdr{}. On WebQ, the $\mathcal{L}_{\text{alt-2}}$ objective diverges and leads to a poor performance. We leave further analysis on the convergence of $\mathcal{L}_{\text{alt-2}}$ objective as a part of future work.


\section{Related Work}
\label{sec:related-work}

Our work is based on end-to-end training of neural readers and retrievers, which we discuss in \S\ref{sec:introduction}, \S\ref{sec:model}, and \S\ref{sec:exp-setup}. Here we instead focus on discussing previous work related to standalone neural retrievers, neural readers, and their application in other natural language processing tasks.

\textbf{Neural retrievers.} \ There are two broad classes of neural retrievers based on the number of embeddings computed for a document: dual encoders~\citep{yih-etal-2011-learning, lee-etal-2019-latent} and multivector encoders~\citep{khattab2020colbert, yi2021sparse}. Dual encoders store one embedding for each evidence document. Multivector encoders require multiple embeddings, which can be computationally expensive for large-scale retrieval. Due to the large size of the evidence document collection in OpenQA, our work uses the more efficient dual-encoder. ~\citet{sachan2021end} show that the performance of supervised dual encoders in OpenQA can be improved when pre-training with the Inverse Cloze Task for the high-resource setting or masked salient spans for the low-resource setting.

\textbf{Neural readers.} \ Neural readers output an answer given retrieved documents as its input. There are also two broad classes of neural readers: extractive and generative. Extractive readers \citep{clark-gardner-2018-simple,mbpa2019,wang-etal-2019-multi,guu2020realm,karpukhin2020dense} extract a span from a retrieved document to produce an answer.
Generative readers \citep{izacard2021leveraging}, on the other hand, generates an answer conditioned on the retrieved documents.

\textbf{Other application areas.} \ In addition to question answering, retrieval-augmented methods have been successfully applied to other natural language processing tasks. In left-to-right language modeling, retrieving similar words from an external memory has been shown to improve perplexity~\citep{Khandelwal2020Generalization,dani2021adaptive}. In machine translation, retrieving domain-specific target language tokens has improved performance in domain adaptation~\citep{khandelwal2021nearest}. Finally, in dialog modeling, retrieving knowledge-informed text has helped improve factual correctness in the generated conversations~\citep{fan2021augmenting}.

We provide a detailed comparison of \emdr{} with some of the previous work in Appendix~\ref{sec:realm-cmp} and \ref{sec:ext-rel-work}.


\section{Discussion}
\label{sec:conclusion}

\paragraph{Summary of contributions.}
We presented \emdr{}, an end-to-end training method for retrieval-augmented question
answering systems. We showed how to arrive at our training objective using the expectation-maximization algorithm. We demonstrated that \emdr{} achieves state-of-the-art performance on three benchmark OpenQA datasets.

\paragraph{Technical limitations.}
\emdr{} shares a few limitations with other retrieval-augmented question answering models. In particular, as evidence documents are stored in an uncompressed format, maintaining them and searching for relevant documents can be expensive (both in terms of compute and memory consumption).
In our experiments, we only focused on open-domain question answering. It would be interesting to see how \emdr{} performs for other text generation models as well. We also note that training is relatively resource-heavy (requiring 16 GPUs), potentially having environmental concerns.

\paragraph{Potential negative societal impacts.}
While \emdr{} has the potential to improve language models in the low-resource setting (as demonstrated by our results on WebQ in \S{\ref{sec:results}}), it could exhibit typical biases that
are associated with large language models. For example, our model does not have an explicit mechanism 
to generate answers that are calibrated for fairness across all spectra. As a retrieval-augmented method, it also could be more prone to generating fake answers if an attacker manages to have access and modify information in the collection of evidence documents.

\clearpage

\section*{Acknowledgements}
The authors would like to thank the DeepMind Language team, Mila's students, and anonymous reviewers for providing us valuable feedback and useful suggestions about this work that helped us improve the paper. 

\section*{Funding Statement}
DSS was supported by the Canada CIFAR AI Chair held by Prof.\ William Hamilton.

\bibliographystyle{natbib}
\bibliography{main.bib}

\begin{thebibliography}{}

\bibitem[Berant {\em et~al.}(2013)Berant, Chou, Frostig, and
  Liang]{berant-etal-2013-semantic}
Berant, J., Chou, A., Frostig, R., and Liang, P. (2013).
\newblock Semantic parsing on {F}reebase from question-answer pairs.
\newblock In {\em Proceedings of the 2013 Conference on Empirical Methods in
  Natural Language Processing\/}.

\bibitem[Bromley {\em et~al.}(1994)Bromley, Guyon, LeCun, S\"{a}ckinger, and
  Shah]{bromley1994signature}
Bromley, J., Guyon, I., LeCun, Y., S\"{a}ckinger, E., and Shah, R. (1994).
\newblock Signature verification using a "siamese" time delay neural network.
\newblock In {\em Advances in Neural Information Processing Systems\/}.

\bibitem[Brown {\em et~al.}(2020)Brown, Mann, Ryder, Subbiah, Kaplan, Dhariwal,
  Neelakantan, Shyam, Sastry, Askell, Agarwal, Herbert-Voss, Krueger, Henighan,
  Child, Ramesh, Ziegler, Wu, Winter, Hesse, Chen, Sigler, Litwin, Gray, Chess,
  Clark, Berner, McCandlish, Radford, Sutskever, and Amodei]{brown2020gpt3}
Brown, T., Mann, B., Ryder, N., Subbiah, M., Kaplan, J.~D., Dhariwal, P.,
  Neelakantan, A., Shyam, P., Sastry, G., Askell, A., Agarwal, S.,
  Herbert-Voss, A., Krueger, G., Henighan, T., Child, R., Ramesh, A., Ziegler,
  D., Wu, J., Winter, C., Hesse, C., Chen, M., Sigler, E., Litwin, M., Gray,
  S., Chess, B., Clark, J., Berner, C., McCandlish, S., Radford, A., Sutskever,
  I., and Amodei, D. (2020).
\newblock Language models are few-shot learners.
\newblock In {\em Advances in Neural Information Processing Systems\/}.

\bibitem[Chen {\em et~al.}(2017)Chen, Fisch, Weston, and
  Bordes]{chen-etal-2017-reading}
Chen, D., Fisch, A., Weston, J., and Bordes, A. (2017).
\newblock Reading {W}ikipedia to answer open-domain questions.
\newblock In {\em Proceedings of the 55th Annual Meeting of the Association for
  Computational Linguistics (Volume 1: Long Papers)\/}.

\bibitem[Clark and Gardner(2018)Clark and Gardner]{clark-gardner-2018-simple}
Clark, C. and Gardner, M. (2018).
\newblock Simple and effective multi-paragraph reading comprehension.
\newblock In {\em Proceedings of the 56th Annual Meeting of the Association for
  Computational Linguistics (Volume 1: Long Papers)\/}.

\bibitem[de~Masson~d'Autume {\em et~al.}(2019)de~Masson~d'Autume, Ruder, Kong,
  and Yogatama]{mbpa2019}
de~Masson~d'Autume, C., Ruder, S., Kong, L., and Yogatama, D. (2019).
\newblock Episodic memory in lifelong language learning.
\newblock In {\em Advances in Neural Information Processing Systems\/}.

\bibitem[Dempster {\em et~al.}(1977)Dempster, Laird, and Rubin]{em}
Dempster, A., Laird, N., and Rubin, D. (1977).
\newblock Maximum likelihood from incomplete data via the em algorithm.
\newblock {\em Journal of the Royal Statistical Society, Series B\/}, {\bf
  1}(39), 1--38.

\bibitem[Devlin {\em et~al.}(2019)Devlin, Chang, Lee, and
  Toutanova]{devlin2019bert}
Devlin, J., Chang, M.-W., Lee, K., and Toutanova, K. (2019).
\newblock {BERT}: Pre-training of deep bidirectional transformers for language
  understanding.
\newblock In {\em Proceedings of the 2019 Conference of the North {A}merican
  Chapter of the Association for Computational Linguistics: Human Language
  Technologies, Volume 1 (Long and Short Papers)\/}.

\bibitem[Fan {\em et~al.}(2021)Fan, Gardent, Braud, and
  Bordes]{fan2021augmenting}
Fan, A., Gardent, C., Braud, C., and Bordes, A. (2021).
\newblock {Augmenting Transformers with KNN-Based Composite Memory for Dialog}.
\newblock {\em Transactions of the Association for Computational
  Linguistics\/}, {\bf 9}.

\bibitem[Guu {\em et~al.}(2020)Guu, Lee, Tung, Pasupat, and
  Chang]{guu2020realm}
Guu, K., Lee, K., Tung, Z., Pasupat, P., and Chang, M. (2020).
\newblock Retrieval augmented language model pre-training.
\newblock In {\em Proceedings of the 37th International Conference on Machine
  Learning\/}.

\bibitem[Iyer {\em et~al.}(2021)Iyer, Min, Mehdad, and Yih]{iyer2020reconsider}
Iyer, S., Min, S., Mehdad, Y., and Yih, W. (2021).
\newblock Reconsider: Re-ranking using span-focused cross-attention for open
  domain question answering.
\newblock In {\em Proceedings of the 2021 Conference of the North {A}merican
  Chapter of the Association for Computational Linguistics: Human Language
  Technologies, Volume 1 (Long and Short Papers)\/}.

\bibitem[Izacard and Grave(2021a)Izacard and Grave]{izacard2021distilling}
Izacard, G. and Grave, E. (2021a).
\newblock Distilling knowledge from reader to retriever for question answering.
\newblock In {\em International Conference on Learning Representations\/}.

\bibitem[Izacard and Grave(2021b)Izacard and Grave]{izacard2021leveraging}
Izacard, G. and Grave, E. (2021b).
\newblock Leveraging passage retrieval with generative models for open domain
  question answering.
\newblock In {\em Proceedings of the 16th Conference of the European Chapter of
  the Association for Computational Linguistics: Main Volume\/}.

\bibitem[Joshi {\em et~al.}(2017)Joshi, Choi, Weld, and
  Zettlemoyer]{joshi2017triviaqa}
Joshi, M., Choi, E., Weld, D., and Zettlemoyer, L. (2017).
\newblock {T}rivia{QA}: A large scale distantly supervised challenge dataset
  for reading comprehension.
\newblock In {\em Proceedings of the 55th Annual Meeting of the Association for
  Computational Linguistics (Volume 1: Long Papers)\/}.

\bibitem[Karpukhin {\em et~al.}(2020)Karpukhin, O{\u{g}}uz, Min, Wu, Edunov,
  Chen, and Yih]{karpukhin2020dense}
Karpukhin, V., O{\u{g}}uz, B., Min, S., Wu, L., Edunov, S., Chen, D., and Yih,
  W.-t. (2020).
\newblock Dense passage retrieval for open-domain question answering.
\newblock In {\em Proceedings of the 2020 Conference on Empirical Methods in
  Natural Language Processing (EMNLP)\/}.

\bibitem[Khandelwal {\em et~al.}(2020)Khandelwal, Levy, Jurafsky, Zettlemoyer,
  and Lewis]{Khandelwal2020Generalization}
Khandelwal, U., Levy, O., Jurafsky, D., Zettlemoyer, L., and Lewis, M. (2020).
\newblock Generalization through memorization: Nearest neighbor language
  models.
\newblock In {\em International Conference on Learning Representations\/}.

\bibitem[Khandelwal {\em et~al.}(2021)Khandelwal, Fan, Jurafsky, Zettlemoyer,
  and Lewis]{khandelwal2021nearest}
Khandelwal, U., Fan, A., Jurafsky, D., Zettlemoyer, L., and Lewis, M. (2021).
\newblock Nearest neighbor machine translation.
\newblock In {\em International Conference on Learning Representations\/}.

\bibitem[Khattab and Zaharia(2020)Khattab and Zaharia]{khattab2020colbert}
Khattab, O. and Zaharia, M. (2020).
\newblock Colbert: Efficient and effective passage search via contextualized
  late interaction over bert.
\newblock In {\em Proceedings of the 43rd International ACM SIGIR Conference on
  Research and Development in Information Retrieval\/}.

\bibitem[Kingma and Ba(2015)Kingma and Ba]{kingma2014adam}
Kingma, D.~P. and Ba, J. (2015).
\newblock Adam: A method for stochastic optimization.
\newblock In {\em The 2015 International Conference for Learning
  Representations\/}.

\bibitem[Kwiatkowski {\em et~al.}(2019)Kwiatkowski, Palomaki, Redfield,
  Collins, Parikh, Alberti, Epstein, Polosukhin, Kelcey, Devlin, Lee,
  Toutanova, Jones, Chang, Dai, Uszkoreit, Le, and
  Petrov]{Kwiatkowski2019natural}
Kwiatkowski, T., Palomaki, J., Redfield, O., Collins, M., Parikh, A., Alberti,
  C., Epstein, D., Polosukhin, I., Kelcey, M., Devlin, J., Lee, K., Toutanova,
  K.~N., Jones, L., Chang, M.-W., Dai, A., Uszkoreit, J., Le, Q., and Petrov,
  S. (2019).
\newblock Natural questions: a benchmark for question answering research.
\newblock {\em Transactions of the Association of Computational Linguistics\/}.

\bibitem[Lee {\em et~al.}(2019)Lee, Chang, and Toutanova]{lee-etal-2019-latent}
Lee, K., Chang, M.-W., and Toutanova, K. (2019).
\newblock Latent retrieval for weakly supervised open domain question
  answering.
\newblock In {\em Proceedings of the 57th Annual Meeting of the Association for
  Computational Linguistics\/}.

\bibitem[Lewis {\em et~al.}(2020a)Lewis, Liu, Goyal, Ghazvininejad, Mohamed,
  Levy, Stoyanov, and Zettlemoyer]{lewis2020bart}
Lewis, M., Liu, Y., Goyal, N., Ghazvininejad, M., Mohamed, A., Levy, O.,
  Stoyanov, V., and Zettlemoyer, L. (2020a).
\newblock {BART}: Denoising sequence-to-sequence pre-training for natural
  language generation, translation, and comprehension.
\newblock In {\em Proceedings of the 58th Annual Meeting of the Association for
  Computational Linguistics\/}.

\bibitem[Lewis {\em et~al.}(2020b)Lewis, Perez, Piktus, Petroni, Karpukhin,
  Goyal, K\"{u}ttler, Lewis, Yih, Rockt\"{a}schel, Riedel, and
  Kiela]{lewis20rag}
Lewis, P., Perez, E., Piktus, A., Petroni, F., Karpukhin, V., Goyal, N.,
  K\"{u}ttler, H., Lewis, M., Yih, W.-t., Rockt\"{a}schel, T., Riedel, S., and
  Kiela, D. (2020b).
\newblock Retrieval-augmented generation for knowledge-intensive nlp tasks.
\newblock In {\em Advances in Neural Information Processing Systems\/}.

\bibitem[Luan {\em et~al.}(2021)Luan, Eisenstein, Toutanova, and
  Collins]{yi2021sparse}
Luan, Y., Eisenstein, J., Toutanova, K., and Collins, M. (2021).
\newblock {Sparse, Dense, and Attentional Representations for Text Retrieval}.
\newblock {\em Transactions of the Association for Computational
  Linguistics\/}, {\bf 9}.

\bibitem[Min {\em et~al.}(2019)Min, Chen, Hajishirzi, and
  Zettlemoyer]{min-etal-2019-discrete}
Min, S., Chen, D., Hajishirzi, H., and Zettlemoyer, L. (2019).
\newblock A discrete hard {EM} approach for weakly supervised question
  answering.
\newblock In {\em Proceedings of the 2019 Conference on Empirical Methods in
  Natural Language Processing and the 9th International Joint Conference on
  Natural Language Processing (EMNLP-IJCNLP)\/}, Hong Kong, China. Association
  for Computational Linguistics.

\bibitem[Paszke {\em et~al.}(2019)Paszke, Gross, Massa, Lerer, Bradbury,
  Chanan, Killeen, Lin, Gimelshein, Antiga, Desmaison, Kopf, Yang, DeVito,
  Raison, Tejani, Chilamkurthy, Steiner, Fang, Bai, and
  Chintala]{paszke2019pytorch}
Paszke, A., Gross, S., Massa, F., Lerer, A., Bradbury, J., Chanan, G., Killeen,
  T., Lin, Z., Gimelshein, N., Antiga, L., Desmaison, A., Kopf, A., Yang, E.,
  DeVito, Z., Raison, M., Tejani, A., Chilamkurthy, S., Steiner, B., Fang, L.,
  Bai, J., and Chintala, S. (2019).
\newblock Pytorch: An imperative style, high-performance deep learning library.
\newblock In H.~Wallach, H.~Larochelle, A.~Beygelzimer, F.~d\textquotesingle
  Alch\'{e}-Buc, E.~Fox, and R.~Garnett, editors, {\em Advances in Neural
  Information Processing Systems\/}.

\bibitem[Petroni {\em et~al.}(2019)Petroni, Rockt{\"a}schel, Riedel, Lewis,
  Bakhtin, Wu, and Miller]{petroni-etal-2019-language}
Petroni, F., Rockt{\"a}schel, T., Riedel, S., Lewis, P., Bakhtin, A., Wu, Y.,
  and Miller, A. (2019).
\newblock Language models as knowledge bases?
\newblock In {\em Proceedings of the 2019 Conference on Empirical Methods in
  Natural Language Processing and the 9th International Joint Conference on
  Natural Language Processing (EMNLP-IJCNLP)\/}.

\bibitem[Qi {\em et~al.}(2020)Qi, Zhang, Zhang, Bolton, and
  Manning]{qi2020stanza}
Qi, P., Zhang, Y., Zhang, Y., Bolton, J., and Manning, C.~D. (2020).
\newblock Stanza: A {Python} natural language processing toolkit for many human
  languages.
\newblock In {\em Proceedings of the 58th Annual Meeting of the Association for
  Computational Linguistics: System Demonstrations\/}.

\bibitem[Raffel {\em et~al.}(2020)Raffel, Shazeer, Roberts, Lee, Narang,
  Matena, Zhou, Li, and Liu]{raffel2020t5}
Raffel, C., Shazeer, N., Roberts, A., Lee, K., Narang, S., Matena, M., Zhou,
  Y., Li, W., and Liu, P.~J. (2020).
\newblock Exploring the limits of transfer learning with a unified text-to-text
  transformer.
\newblock {\em Journal of Machine Learning Research\/}, {\bf 21}(140), 1--67.

\bibitem[Roberts {\em et~al.}(2020)Roberts, Raffel, and
  Shazeer]{roberts-etal-2020-much}
Roberts, A., Raffel, C., and Shazeer, N. (2020).
\newblock How much knowledge can you pack into the parameters of a language
  model?
\newblock In {\em Proceedings of the 2020 Conference on Empirical Methods in
  Natural Language Processing (EMNLP)\/}.

\bibitem[Robertson and Zaragoza(2009)Robertson and Zaragoza]{Robertson2009bm25}
Robertson, S. and Zaragoza, H. (2009).
\newblock The probabilistic relevance framework: Bm25 and beyond.
\newblock {\em Foundations and Trends in Information Retrieval\/}.

\bibitem[Sachan {\em et~al.}(2021)Sachan, Patwary, Shoeybi, Kant, Ping,
  Hamilton, and Catanzaro]{sachan2021end}
Sachan, D.~S., Patwary, M., Shoeybi, M., Kant, N., Ping, W., Hamilton, W.~L.,
  and Catanzaro, B. (2021).
\newblock End-to-end training of neural retrievers for open-domain question
  answering.
\newblock In {\em Joint Conference of the 59th Annual Meeting of the
  Association for Computational Linguistics and the 11th International Joint
  Conference on Natural Language Processing (ACL-IJCNLP)\/}.

\bibitem[Shoeybi {\em et~al.}(2019)Shoeybi, Patwary, Puri, LeGresley, Casper,
  and Catanzaro]{shoeybi2019megatron}
Shoeybi, M., Patwary, M., Puri, R., LeGresley, P., Casper, J., and Catanzaro,
  B. (2019).
\newblock Megatron-lm: Training multi-billion parameter language models using
  gpu model parallelism.
\newblock {\em arXiv preprint arXiv:1909.08053\/}.

\bibitem[Vaswani {\em et~al.}(2017)Vaswani, Shazeer, Parmar, Uszkoreit, Jones,
  Gomez, Kaiser, and Polosukhin]{vaswani2017attention}
Vaswani, A., Shazeer, N., Parmar, N., Uszkoreit, J., Jones, L., Gomez, A.~N.,
  Kaiser, {\L}., and Polosukhin, I. (2017).
\newblock Attention is all you need.
\newblock In {\em Advances in Neural Information Processing Systems\/}.

\bibitem[Wang {\em et~al.}(2018)Wang, Yu, Guo, Wang, Klinger, Zhang, Chang,
  Tesauro, Zhou, and Jiang]{DBLP:conf/aaai/WangYGWKZCTZJ18}
Wang, S., Yu, M., Guo, X., Wang, Z., Klinger, T., Zhang, W., Chang, S.,
  Tesauro, G., Zhou, B., and Jiang, J. (2018).
\newblock R3: Reinforced ranker-reader for open-domain question answering.
\newblock In {\em AAAI\/}.

\bibitem[Wang {\em et~al.}(2019)Wang, Ng, Ma, Nallapati, and
  Xiang]{wang-etal-2019-multi}
Wang, Z., Ng, P., Ma, X., Nallapati, R., and Xiang, B. (2019).
\newblock Multi-passage {BERT}: A globally normalized {BERT} model for
  open-domain question answering.
\newblock In {\em Proceedings of the 2019 Conference on Empirical Methods in
  Natural Language Processing and the 9th International Joint Conference on
  Natural Language Processing (EMNLP-IJCNLP)\/}.

\bibitem[Yih {\em et~al.}(2011)Yih, Toutanova, Platt, and
  Meek]{yih-etal-2011-learning}
Yih, W.-t., Toutanova, K., Platt, J.~C., and Meek, C. (2011).
\newblock Learning discriminative projections for text similarity measures.
\newblock In {\em Proceedings of the Fifteenth Conference on Computational
  Natural Language Learning\/}.

\bibitem[Yogatama {\em et~al.}(2021)Yogatama, de~Masson~d’Autume, and
  Kong]{dani2021adaptive}
Yogatama, D., de~Masson~d’Autume, C., and Kong, L. (2021).
\newblock {Adaptive Semiparametric Language Models}.
\newblock {\em Transactions of the Association for Computational
  Linguistics\/}, {\bf 9}, 362--373.

\end{thebibliography}

\clearpage

\section*{Checklist}


\begin{enumerate}

\item For all authors...
\begin{enumerate}
  \item Do the main claims made in the abstract and introduction accurately reflect the paper's contributions and scope?
    \answerYes{Please see the model (\S\ref{sec:model}) and result (\S\ref{sec:exp-setup}) sections that solidify the claims made in the abstract and introduction sections.}
  \item Did you describe the limitations of your work?
    \answerYes{Please see limitations in \S\ref{sec:conclusion}.}
  \item Did you discuss any potential negative societal impacts of your work?
    \answerYes{Please see negative societal impact in \S\ref{sec:conclusion}.}
  \item Have you read the ethics review guidelines and ensured that your paper conforms to them?
    \answerYes{}
\end{enumerate}

\item If you are including theoretical results...
\begin{enumerate}
  \item Did you state the full set of assumptions of all theoretical results?
    \answerNA{}
	\item Did you include complete proofs of all theoretical results?
    \answerNA{}
\end{enumerate}

\item If you ran experiments...
\begin{enumerate}
  \item Did you include the code, data, and instructions needed to reproduce the main experimental results (either in the supplemental material or as a URL)?
    \answerYes{We include the code, data, and instructions in the supplemental material and \S{\ref{sec:impl-details}}.}
  \item Did you specify all the training details (e.g., data splits, hyperparameters, how they were chosen)?
    \answerYes{We specify these details in the appendix included in the supplementary material.}
	\item Did you report error bars (e.g., with respect to the random seed after running experiments multiple times)?
    \answerNo{Our experiments are compute expensive and it is not feasible to perform multiple runs of the same experiment with different seeds. All our training runs 
    use the same seed value (1234).    
    As an alternative to running multiple seeds, we perform a number of ablation studies (\S{\ref{sec:ablation}}).}
	\item Did you include the total amount of compute and the type of resources used (e.g., type of GPUs, internal cluster, or cloud provider)?
    \answerYes{Please see \S\ref{sec:impl-details} under hardware and library.}
\end{enumerate}

\item If you are using existing assets (e.g., code, data, models) or curating/releasing new assets...
\begin{enumerate}
  \item If your work uses existing assets, did you cite the creators?
    \answerYes{Please see \S\ref{sec:dataset} for the details.}
  \item Did you mention the license of the assets?
    \answerYes{Our work is based on open-source data and framework. When applicable, we describe the license information in the appendix.}
  \item Did you include any new assets either in the supplemental material or as a URL?
    \answerYes{} We include our code in the supplementary material.
  \item Did you discuss whether and how consent was obtained from people whose data you're using/curating?
    \answerNA{}
  \item Did you discuss whether the data you are using/curating contains personally identifiable information or offensive content?
    \answerNA{}
\end{enumerate}

\item If you used crowdsourcing or conducted research with human subjects...
\begin{enumerate}
  \item Did you include the full text of instructions given to participants and screenshots, if applicable?
    \answerNA{}
  \item Did you describe any potential participant risks, with links to Institutional Review Board (IRB) approvals, if applicable?
    \answerNA{}
  \item Did you include the estimated hourly wage paid to participants and the total amount spent on participant compensation?
    \answerNA{}
\end{enumerate}

\end{enumerate}


\clearpage
\appendix

\section{Dataset Details} \label{appendix:dataset-stat}

\paragraph*{Dataset statistics.} For validation, we randomly select approximately 10\% examples from the training set. For all the datasets, we use the dataset splits from~\citep{lee-etal-2019-latent}. We provide the size of the training, development, and test sets in Table~\ref{table:dataset-stat}.

\paragraph*{Pre-processing.} For TriviaQA experiments, following~\citep{izacard2021distilling}, we select human-annotated answers for training the QA model. We also filter out those questions whose answer length is more than $5$ words. Overall, this filters out 2,362 examples from the training set.

\begin{table}[t]
\small
\centering
\begin{tabular}{l c c c c}
  \toprule
  \textbf{Dataset} & \textbf{Train} & \textbf{Filtered Train} & \textbf{Dev} & \textbf{Test} \\
  \midrule
  WebQuestions (WebQ) & \phantom{0}3,417 & \phantom{0}2,474 & \phantom{00}361 & \phantom{0}2,032 \\
  Natural Questions (NQ) & 79,168 & 58,880 & 8,757 & \phantom{0}3,610 \\
  TriviaQA & 78,785 & 60,413 & 8,837 & 11,313 \\
  \bottomrule
\end{tabular}
\vspace{2mm}
\caption{OpenQA dataset statistics. The training set is used for end-to-end training of the QA models whereas the filtered training set is used for supervised training of the retriever (i.e., for DPR experiments). The filtered set ignores those question-answer pairs where the evidence (Wikipedia) document retrieved using BM25~\citep{Robertson2009bm25} does not align with the provided gold context documents. We leverage the filtered training set as provided by~\citep{karpukhin2020dense}.}
\label{table:dataset-stat}
\end{table}

\paragraph{Dataset license and URLs.} All the datasets are open-source and widely used by the community. Below, we provide the URLs of the actual dataset source and their preprocessed version which is used in this work.
\begin{itemize} 
	\item NQ: \textit{dataset}: \url{https://ai.google.com/research/NaturalQuestions/download}, \textit{license}: \url{https://github.com/google-research-datasets/natural-questions/blob/master/LICENSE}
	\item TriviaQA: \textit{dataset}: \url{http://nlp.cs.washington.edu/triviaqa/}, \textit{license}: \url{https://github.com/mandarjoshi90/triviaqa/blob/master/LICENSE}
	\item WebQ: \textit{dataset}: \url{https://github.com/google-research/language/tree/master/language/orqa#getting-the-data}, \textit{license}: \url{https://nlp.stanford.edu/software/sempre/}
	\item Preprocessed version: We make use of NQ, TriviaQA, and evidence datasets as open-sourced by~\cite{karpukhin2020dense} here: \url{https://github.com/facebookresearch/DPR/blob/master/data/download_data.py}.
\end{itemize}

\section{Additional Training Details}

\begin{table*}[t]
\small
\centering
\begin{tabular}{@{}l c c c c@{}}
 \toprule 
 \textbf{Hyperparameter} & \textbf{BERT} & \textbf{ICT} & \textbf{T5} & \textbf{MSS} \\
 \midrule
  Dataset & Wikipedia, BookCorpus & Wikipedia & C4, Wikipedia, OpenWebText & Wikipedia \\
  Num.\ Parameters & 110M & 220M & 220M & 440M \\
  Hidden Size & 768 & 768 & 768 & 768\\
  Attention heads & 12 & 12 & 12 & 12 \\
  Dropout & 0.1 & 0.1 & 0.1 & 0.1 \\
  Optimizer & Adam & Adam & Adam & Adam \\
  Batch Size & 256 & 4096 & 2048 & 64 \\
  Training Steps & 1M & 100K & 1M & 82K \\
  Warmup Ratio & 0.01 & 0.01 & 0.01 & 0.05 \\
  Max.\ Learning Rate & 1e-4 & 1e-4 & 1e-4 & 2e-5 \\
  Weight Decay & 1e-2 & 1e-2 & 1e-2 & 1e-1 \\
  Learning Rate Decay & Linear & Linear & Linear & Linear \\
  Gradient Clipping & 1.0 & 1.0 & 1.0 & 1.0	\\
  \bottomrule
\end{tabular}
\caption{Hyperparameters for training BERT, ICT, T5, and MSS models.}
\label{tab:hyperparams}
\end{table*}

\begin{table*}[t]
\small
\centering
\begin{tabular}{l c c c}
 \toprule 
 \textbf{Hyperparameter} & \textbf{NQ} & \textbf{TriviaQA} & \textbf{WebQ} \\
 \midrule
  Num.\ Parameters & 440M & 440M & 440M \\
  Hidden Size & 768 & 768 & 768\\
  Attention heads & 12 & 12 & 12 \\
  Dropout & 0.1 & 0.1 & 0.1 \\
  Optimizer & Adam & Adam & Adam \\
  Batch Size & 64 & 64 & 16 \\
  Epochs & 10 & 10 & 20 \\
  Warmup Ratio & 0.01 & 0.01 & 0.01 \\
  Max.\ Learning Rate & 2e-5 & 2e-5 & 2e-5 \\
  Weight Decay & 1e-1 & 1e-1 & 1e-1 \\
  Learning Rate Decay & Linear & Linear & Linear \\
  Gradient Clipping & 1.0 & 1.0 & 1.0 	\\
  Temperate ($\tau$) & 27.7 & 27.7 & 27.7 \\
  \bottomrule
\end{tabular}
\caption{Hyperparameters for finetuning on NQ, TriviaQA, and WebQ datasets.}
\label{tab:sup-hyperparams}
\end{table*}

In addition to the details provided in \S\ref{sec:impl-details}, here, we provide further training details for reproducibility.

\paragraph{BERT and Inverse Cloze Task (ICT).} We derive the implementations of BERT~\citep{devlin2019bert} and ICT~\citep{lee-etal-2019-latent} from the open-source Megatron-LM toolkit.\footnote{\url{https://github.com/NVIDIA/Megatron-LM}}
For ICT, the dual-encoder retriever is initialized with BERT weights and then we train the model according to~\cite{lee-etal-2019-latent}. 
For training, we use Wikipedia paragraphs where we truncate the maximum length of a paragraph to 256 tokens.
We list the settings and hyperparameters used for training BERT and ICT in Table~\ref{tab:hyperparams}.

\paragraph{T5.} We derive the implementation of T5~\citep{raffel2020t5} language model from the open-source Megatron-LM toolkit~\citep{shoeybi2019megatron}. 
We list the hyperparameters used for training T5 in Table~\ref{tab:hyperparams}. For consistency, we train T5 for the same number of steps and batch size as was done in the original paper. 
Additionally, we use BERT lowercase tokenization for both T5 and BERT.

\paragraph*{Unsupervised pre-training with masked salient spans (MSS).}
For MSS training, we initialize the retriever of our model from the ICT weights and the reader from the T5 weights. 
We make use of the Stanza toolkit~\citep{qi2020stanza} to segment evidence documents into sentences. We then extract named entities from these sentences using the NER model trained on the OntoNotes-5.0 dataset as provided by Stanza. These names entities are replaced by mask tokens. As the masked tokens correspond to special named entities, they are referred to as salient spans. 
The masked sentence is considered as the question to retrieve evidence documents and the reader is trained to generate the named entities corresponding to the masked salient spans with the help of retrieved documents. During retrieval, we ignore the evidence document from which the masked sentence was derived. 
We list the hyperparameters of MSS training in Table~\ref{tab:hyperparams}.

\paragraph*{Supervised training using the question-answer pairs.}
We provide the training details in \S\ref{sec:impl-details}. 
We list the hyperparameters in Table~\ref{tab:sup-hyperparams}.
Apart from the number of epochs and batch size in WebQ, we use the same hyperparameters for all the experiments.
For the temperature parameter ($\tau$) in Eq.~\ref{eq:ret-prob}, we follow~\cite{sachan2021end} and set it as the square root of the hidden size.

\paragraph{Training Time.}
We run all of our experiments on a machine with 96 CPUs, 1.3TB physical memory, and 16 A100 GPUs.
We use PyTorch \citep{paszke2019pytorch} to implement our proposed model.
With this hardware setup, our experiments on NQ and TriviaQA took approximately 25 hours to complete, while experiments on WebQ took roughly 8 hours to complete. Before supervised training, we also perform a one-time unsupervised MSS pre-training for 82,000 steps that took roughly 1 week. 

\section{Unsupervised Pre-training and Comparisons with REALM} \label{sec:realm-cmp}

We make use of a couple of training techniques introduced in the REALM paper~\citep{guu2020realm}: masked salient spans (MSS) pre-training and asynchronous evidence embedding update. There are similarities and differences in the way in which we apply these ideas to \emdr{} training.

\subsection{ICT and MSS Pre-training}

Both ICT and MSS are unsupervised techniques used to bootstrap the retriever so that it has a good initial recall.

We first initialize the retriever with ICT pre-training. For ICT, similar to REALM, we follow the settings in the ORQA paper~\citep{lee-etal-2019-latent}. We observe our Recall@5 to be much higher than that reported in the REALM paper (see Table~\ref{tab:cmp-with-realm}). We believe that our choice of 768 dimensional embedding of each evidence document leads to better results when compared to the 128 dimensional embedding used in REALM.

We further pre-train with MSS once the retriever weights are initialized with ICT. We use a batch size of 64 and train for 82K steps using the \emdr{} objective. In comparison, REALM uses a batch size of 512 and trains the model for 200K steps. 
Even with a much smaller batch size and training steps, \emdr{} achieves similar Recall@5 after MSS training (Table~\ref{tab:cmp-with-realm}). 
We hypothesize that with a large batch size and longer training, \emdr{} would be able to further improve its recall. Another implementation detail is that \emdr{} does not require the additional null document which was used in REALM.

\begin{table}
\small
\centering
\begin{tabular}{l c c}
\toprule
\textbf{Method} & \textbf{R@5 after ICT} & \textbf{R@5 after MSS} \\
\midrule
REALM~\citep{guu2020realm} & 13.9 & 38.5 \\
\emdr{arg} & 28.0 & 38.6 \\
\bottomrule
\end{tabular}
\vspace{2mm}
\caption{Retrieval recall on the NQ development set after ICT and MSS pre-training.}
\label{tab:cmp-with-realm}
\end{table}

For low-resource datasets such as WebQ, MSS pre-training also improves the performance of the FiD reader. As Table~\ref{tab:ret-init} illustrates, on WebQ, MSS pre-trained reader obtains a gain of more than 1 EM point over the T5 reader (shaded in orange color).

\subsection{Asynchronous Evidence Embedding Updates}

The asynchronous evidence embedding updates are performed after every 500 steps of training and is similar to REALM with a couple of differences. In our work, asynchronous embedding updates is done both during MSS pre-training and supervised training, while in REALM it is performed only during MSS pre-training. The second difference, although a minor one, we needed to compute the embeddings of 21M evidence documents while REALM had to do this for 13M documents. We do this by having two process groups during training, one group trains the model on 8 GPUs while the other group performs evidence embedding computation on 8 GPUs in an asynchronous manner.

\subsection{Pre-computed Evidence Embeddings Storage for Retrieval}

\begin{table}
\small
\centering
\begin{tabular}{l c c c}
\toprule
\textbf{Method} & \textbf{Evidence Size} & \textbf{Evidence Dimension} & \textbf{GPU Memory (in FP16)} \\
\midrule
REALM~\citep{guu2020realm} & 13M & 128 & \phantom{0}3 GB \\
\emdr{arg} & 21M & 768 & 30 GB \\
\bottomrule
\end{tabular}
\vspace{2mm}
\caption{Comparison of evidence embeddings storage for retrieval.}
\label{tab:cmp-evidence}
\end{table}

In Table~\ref{tab:cmp-evidence}, we provide some comparisons between REALM and \emdr{} to showcase that the retrieval task is more challenging in our setting. 
Firstly, the size of evidence in REALM is 13M because each Wikipedia article is split into 288 wordpieces while the
size of evidence in \emdr{} is 21M as each Wikipedia article is split into 100 linguistic words.
Second, the embedding dimension of each evidence document in REALM is 128 while the embedding dimension of each evidence document in \emdr{} is 768.
Due to these factors, the memory required by REALM to store evidence embeddings (in FP16) is approximately 3 GB, while the memory required by \emdr{} to store evidence embeddings (in FP16) is 30 GB.
As the GPU RAM is constrained by its capacity (40 GB maximum in A100 GPUs), it was not possible to store the entire 30 GB embeddings in each GPU. Therefore, for online retrieval, we store the evidence embeddings in a distributed fashion over 16 GPUs and perform distributed asynchronous MIPS for fast retrieval.

\section{Comparison with Previous Work} \label{sec:ext-rel-work}
Here we provide a discussion of how \emdr{} is different from some of the previous work.

\subsection{Comparison with Hard EM and Reinforced Reader-Ranker Models}

There are some similarities between \emdr{} and $\mathcal{L}_\text{alt-2}$ to Hard EM~\citep{min-etal-2019-discrete} and Reinforced Reader-Ranker ($\text{R}^3$; \cite{DBLP:conf/aaai/WangYGWKZCTZJ18}), at the conceptual level even though they are not equivalent. Training with REINFORCE involves sampling from a policy network (i.e., the retriever in our case). We take a deterministic approach and take the top-K documents in both \emdr{} and $\mathcal{L}_\text{alt-2}$. Compared to Hard EM, $\mathcal{L}_\text{alt-2}$ directly minimizes the KL divergence of the probability of a retrieved document with the probability of an answer given that document.

At the implementation level, there are many other differences between $\mathcal{L}_\text{alt-2}$ (and \emdr{}) with models in~\citep{min-etal-2019-discrete} and~\citep{DBLP:conf/aaai/WangYGWKZCTZJ18}. First, we would like to note that both these methods use TF-IDF and BM25 as their retrieval approach which are not trainable. In contrast, our work uses a dense retriever which is trained in an end-to-end manner. We list other differences in more detail below.

\paragraph*{Differences with Hard EM.}
~\cite{min-etal-2019-discrete} propose a hard EM approach to train an extractive reader model for QA tasks. The context document is assumed to contain multiple mentions of the correct answer. They propose an objective to train the reader. Specifically, during the training step, the model is trained using maximum marginal likelihood for the first $\tau$ steps and subsequently with their proposed logmax objective. In their open-domain QA experiments on TriviaQA and NQ, the retriever part is based on TF-IDF and BM25 and is non-trainable. Overall, their model is applicable to extractive readers without retriever training. In comparison, in \emdr{}, we train both the reader and retriever. As such, the hard EM approach is not directly applicable to our case.

\paragraph*{Differences with $\text{R}^3$.}
This paper involves three pipelined components: retriever, ranker, and reader. The retriever is BM25 based and is non-trainable. They jointly train the ranker and the reader. The ranker takes 100 documents from the retriever and selects one document to give as input to the reader (contrast this with our work that selects a set of documents). As this selection operation is non-differentiable, their model leverages policy gradient to train the ranker. They also propose a custom reward function based on the overlap of text between the extracted answer and the correct answer. The reader takes a single document as input. In contrast, our approach does not involve a ranker component, both the FiD reader and retriever are trainable, and our proposed objective function \emdr{} is end-to-end differentiable.

\subsection{Comparison with Individual Top-K and Joint Top-K Models}

\paragraph{Comparison with Individual Top-K~\citep{sachan2021end}.}
Individual Top-K is another approach for end-to-end training but the difference is that it applies a single-document reader while \emdr{} consists of a multi-document reader. Similar to previous methods like REALM and RAG, Individual Top-K objective function is also defined over multiple retrieved documents but is better optimized than them. As the performance of \emdr{} is much better than Individual Top-K, \emdr{} is a better modeling approach.

\paragraph{Comparison with Joint Top-K~\citep{sachan2021end}.}
While both \emdr{} and Joint Top-K are end-to-end training approaches for open-domain QA based on the FiD model, they are different in many ways. (i) \emph{Different Objective Functions}: These approaches optimize different training objectives. To achieve retriever training, Joint Top-K adds the retrieval probability score of the top-K documents to the unnormalized inter-attention scores. In this way, the reader pays more importance to those top-K documents with a higher retriever score. There is no explicit feedback from the reader to the retriever. In contrast, the second term in the training objective of \emdr{} explicitly encourages the retriever to improve its predictions based on the agreement with the reader’s answer-generation likelihood of a particular top-K document.
(ii) \emph{Task Performance}: \emdr{} objective leads to a much improved end-to-end training algorithm. This is reflected by the performance gains over the FiD baseline. On NQ and TriviaQA, while \emdr{} leads to 4.3 and 6.4 EM points improvements respectively, Joint Top-K obtains a much lower gain of 1 point improvement on NQ and no improvements on TriviaQA. This demonstrates that EMDR2 training leads to substantially better retrieval, that in turn leads to higher gains in answer generation. These results also illustrate that \emdr{} is a much better end-to-end or joint training algorithm than Joint Top-K for the multi-document reader retriever approaches.

\section{Qualitative Analysis}
\label{sec:qual-analysis}
In Table~\ref{table:examples}, we present some representative examples of the retriever output with both MSS pre-training and when the MSS pre-trained model is finetuned on NQ. 
We observe that after MSS pre-training, the top-$1$ outputs are related to the question but are not relevant enough to answer them.
However, when the MSS pre-trained model is finetuned on NQ with \emdr{}, the retrieval accuracy improves with the top-$1$ documents being much more relevant to answer the question. The retriever's confidence score of the top-$1$ document also improves.

\begin{table*}[t]
\begin{minipage}{\linewidth}
\small
\centering
\begin{tabularx}{\linewidth}{@{} X l X X @{}}
\toprule
\textbf{Questions from NQ test} & \textbf{Answer} & \textbf{MSS Pre-training} & \textbf{\emdr{} finetuned on NQ} \\
\midrule
what type of reaction occurs to form a dipeptide & peptide bond & \textcolor{orange}{probability=0.39} \ldots Bornyl diphosphate synthase In enzymology, bornyl diphosphate synthase (BPPS) () is an enzyme that catalyzes the chemical reaction Bornyl diphosphate synthase is involved in the biosynthesis of the cyclic monoterpenoid bornyl diphosphate. As seen from the reaction above, BPPS takes geranyl diphosphate as its only substrate and isomerizes into the product, (+)- bornyl diphosphate. This reaction comes from a general class of enzymes called terpene synthases that \ldots & 
\textcolor{orange}{probability=0.78} \ldots Subsequent to this coupling reaction, the amine protecting group P and the ester are converted to the free amine and carboxylic acid, respectively. For many amino acids, the ancillary functional groups are protected. The condensation of the amine and the carboxylic acid to form the \textcolor{blue}{peptide bond} generally employs coupling agents to activate the carboxylic acid. The Bergmann azlactone peptide synthesis is a classic organic synthesis for the preparation of dipeptides. \ldots \\
\midrule
when was the japanese videogame company nintendo founded & 23 September 1889 & 
\textcolor{orange}{probability=0.37} \ldots contributed to the development of the following games. Creatures (company) Ape, Inc. was founded in March 1989 and Shigesato Itoi became its chief executive officer. Nintendo president Hiroshi Yamauchi had wanted to support new talent in game design. Liking Itoi\'s work, he proposed the idea of the company to Itoi and invested in it. Ape\'s staff included Tsunekazu Ishihara, who later became the Pokémon Company\'s CEO, and Ashura Benimaru Itoh, a renowned illustrator. They began work on "Mother", which released in July. Its music was composed by Hip Tanaka, who later became the second CEO of Creatures \ldots & 
\textcolor{orange}{probability=0.61} \ldots Nintendo Co., Ltd. is a Japanese multinational consumer electronics and video game company headquartered in Kyoto. Nintendo is one of the world's largest video game companies by market capitalisation, creating some of the best-known and top-selling video game franchises, such as ``Mario'', ``The Legend of Zelda'', and ``Pokémon''. Founded on \textcolor{blue}{23 September 1889} by Fusajiro Yamauchi, it originally produced handmade hanafuda playing cards. By 1963, the company had tried several small niche businesses, such as cab services and love hotels. Abandoning previous ventures in favour of toys in the 1960s \ldots \\
\bottomrule
\end{tabularx}
\caption{Examples of top-1 retrieved documents from the NQ test when the model is pre-trained with Masked Salient Spans (MSS) or finetuned on NQ data. If the answer exists in the document it is highlighted in \textcolor{blue}{blue} color, and the probability of the document (Eq.~\ref{eq:ret-prob}) is indicated in \textcolor{orange}{orange} color.}
\label{table:examples}
\end{minipage}
\end{table*}

\begin{figure*}[!t]
\centering
\includegraphics[max width=.85\textwidth, scale=0.90]{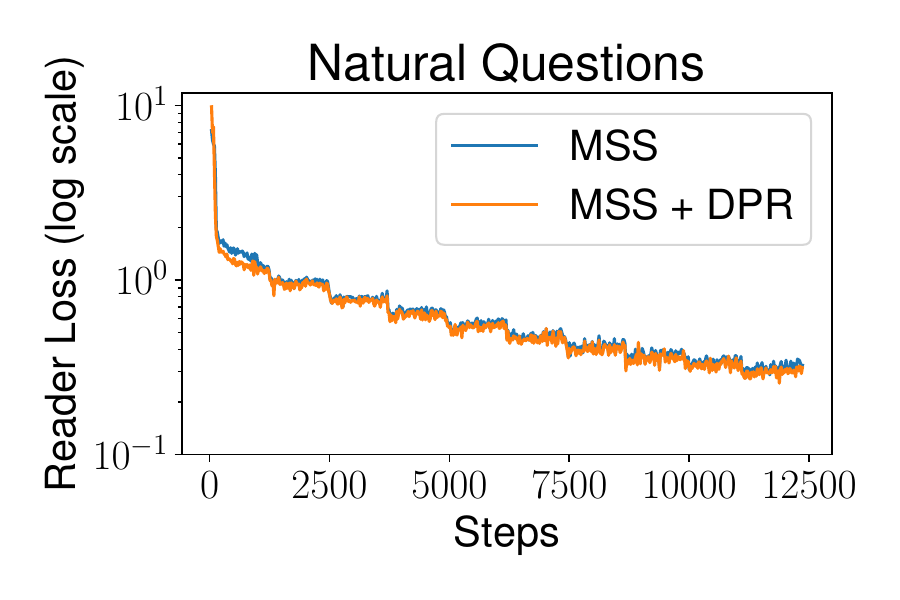}
\includegraphics[max width=.85\textwidth, scale=0.90]{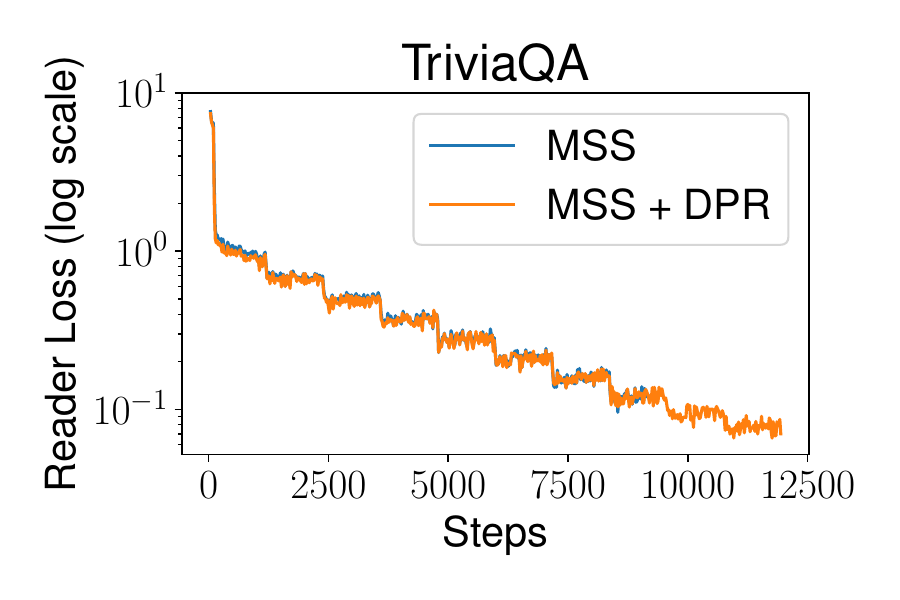}
\includegraphics[max width=.85\textwidth, scale=0.90]{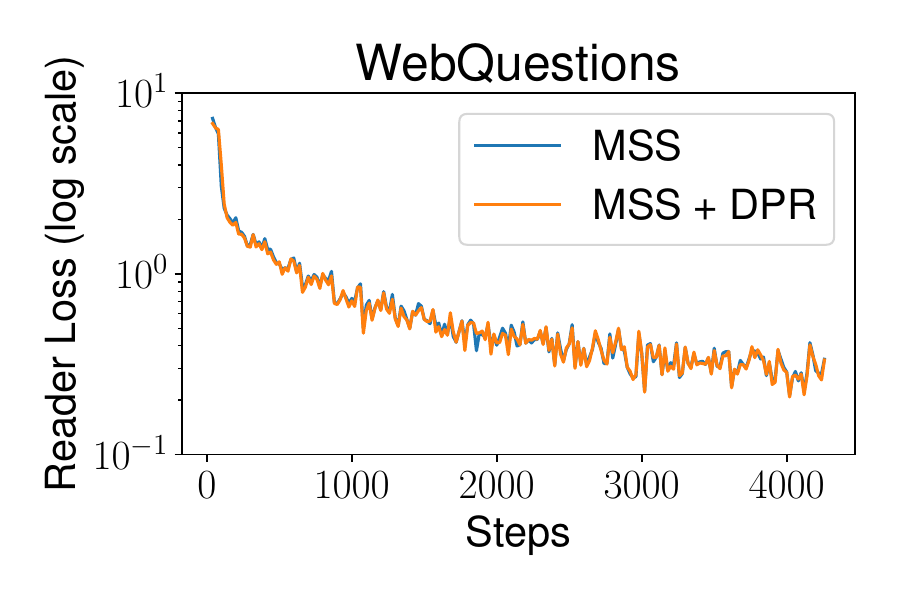}
\vspace{-6pt}
\caption{Reader training loss vs steps for NQ, TriviaQA, and WebQ when the retriever is either initialized by MSS pre-training or by MSS followed by supervised DPR training (MSS + DPR).}
\label{fig:train-loss}
\vspace{-6pt}
\end{figure*}

\paragraph*{Comparing retriever initializations.} 
We analyze the reader's training loss when the retriever is either initialized with unsupervised MSS training or with first MSS pre-training followed by supervised DPR training (MSS + DPR). As indicated in Table~\ref{tab:ret-init}, MSS pre-training being unsupervised has a lower accuracy while MSS + DPR retriever has a higher accuracy.
However, as is also evident from the plots in Figure~\ref{fig:train-loss}, retriever initialization has a marginal effect on the answer generation performance. 
We see that for NQ, for the first 1200 steps, the higher accuracy MSS + DPR retriever leads to a smaller training loss compared with the MSS retriever, after which the difference between the two training losses diminishes as the end-to-end training improves the accuracy of the MSS retriever. Similar trends are also observed for TriviaQA and WebQ but to a lesser extent.

\begin{figure*}[!t]
\centering
\includegraphics[max width=.85\textwidth, scale=0.90]{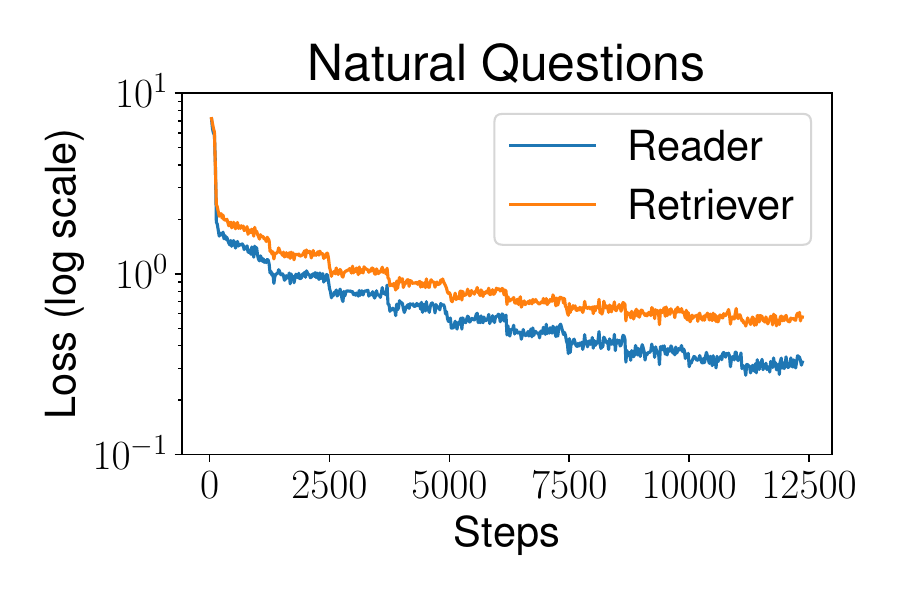}
\includegraphics[max width=.85\textwidth, scale=0.90]{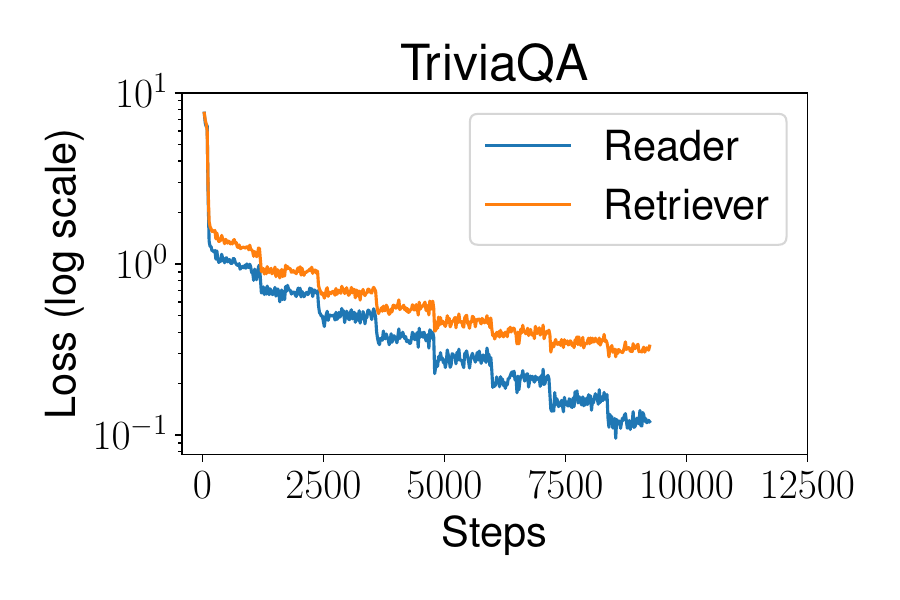}
\includegraphics[max width=.85\textwidth, scale=0.90]{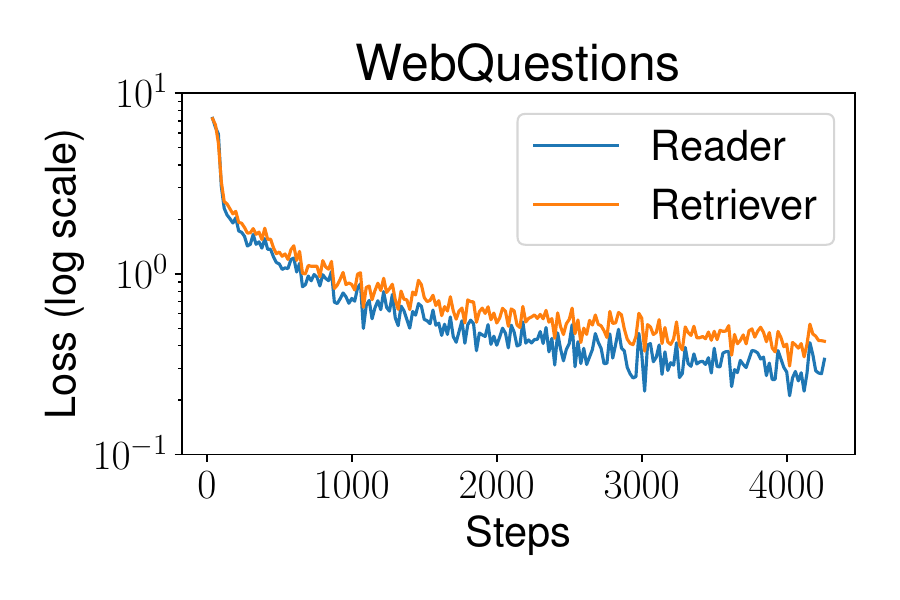}
\vspace{-6pt}
\caption{Reader and retriever training losses when the model is initialized with MSS pre-training.}
\label{fig:reader-retriever-loss}
\vspace{-6pt}
\end{figure*}

\paragraph*{Visualizing reader and retriever losses.} In Figure~\ref{fig:reader-retriever-loss}, we show the trajectories of the reader and retriever training losses when the model is initialized with MSS pre-training.

\end{document}